\documentclass[10pt,twocolumn,letterpaper]{article}

\usepackage[pagenumbers]{cvpr} % To force page numbers, e.g. for an arXiv version

\definecolor{cvprblue}{rgb}{0.21,0.49,0.74}
\usepackage[pagebackref,breaklinks,colorlinks,allcolors=cvprblue]{hyperref}
\usepackage{multirow}
\usepackage{subcaption}
\usepackage{array}
\usepackage{dcolumn}

\def\paperID{5754} % *** Enter the Paper ID here
\def\confName{CVPR}
\def\confYear{2025}

\title{\textit{What You See Is What Matters:} A Novel Visual and Physics-Based Metric for Evaluating Video Generation Quality}

\author{
\begin{tabular}{c}
Zihan Wang \textsuperscript{*}\textsuperscript{\textdagger} \and Songlin Li\textsuperscript{*} \and Lingyan Hao \and Xinyu Hu \and Bowen Song
\end{tabular} \\
Stanford University \\
{\tt\small \{wangzih, svli97, lingyanh, xhu17, bowens18\}@stanford.edu}
}

\begin{document}
\maketitle
\renewcommand{\thefootnote}
{\fnsymbol{footnote}}
\footnotetext[1]{Equal contribution.}
\footnotetext[2]{Project lead and corresponding author.}

\begin{abstract}
As video generation models advance rapidly, assessing the quality of generated videos has become increasingly critical. Existing metrics, such as Fréchet Video Distance (FVD), Inception Score (IS), and ClipSim, measure quality primarily in latent space rather than from a human visual perspective, often overlooking key aspects like appearance and motion consistency to physical laws. In this paper, we propose a novel metric, \textbf{\textit{VAMP}} (\underline{V}isual \underline{A}ppearance and \underline{M}otion \underline{P}lausibility), that evaluates both the visual appearance and physical plausibility of generated videos. VAMP is composed of two main components: an appearance score, which assesses color, shape, and texture consistency across frames, and a motion score, which evaluates the realism of object movements. We validate VAMP through two experiments: corrupted video evaluation and generated video evaluation. In the corrupted video evaluation, we introduce various types of corruptions into real videos and measure the correlation between corruption severity and VAMP scores. In the generated video evaluation, we use state-of-the-art models to generate videos from carefully designed prompts and use VAMP to compare the models' performances.
% and compare VAMP's performance to human evaluators' rankings. 
Our results demonstrate that VAMP effectively captures both visual fidelity and temporal consistency, offering a more comprehensive evaluation of video quality than traditional methods. 
%You can find more information on our \textcolor{orange}{\href{https://www.example.com}{website}}.
\end{abstract}    
\section{Introduction}
\begin{figure}[t]
    \centering
    % Uncomment the following line when you have the actual image file
    \includegraphics[width=0.95\linewidth]{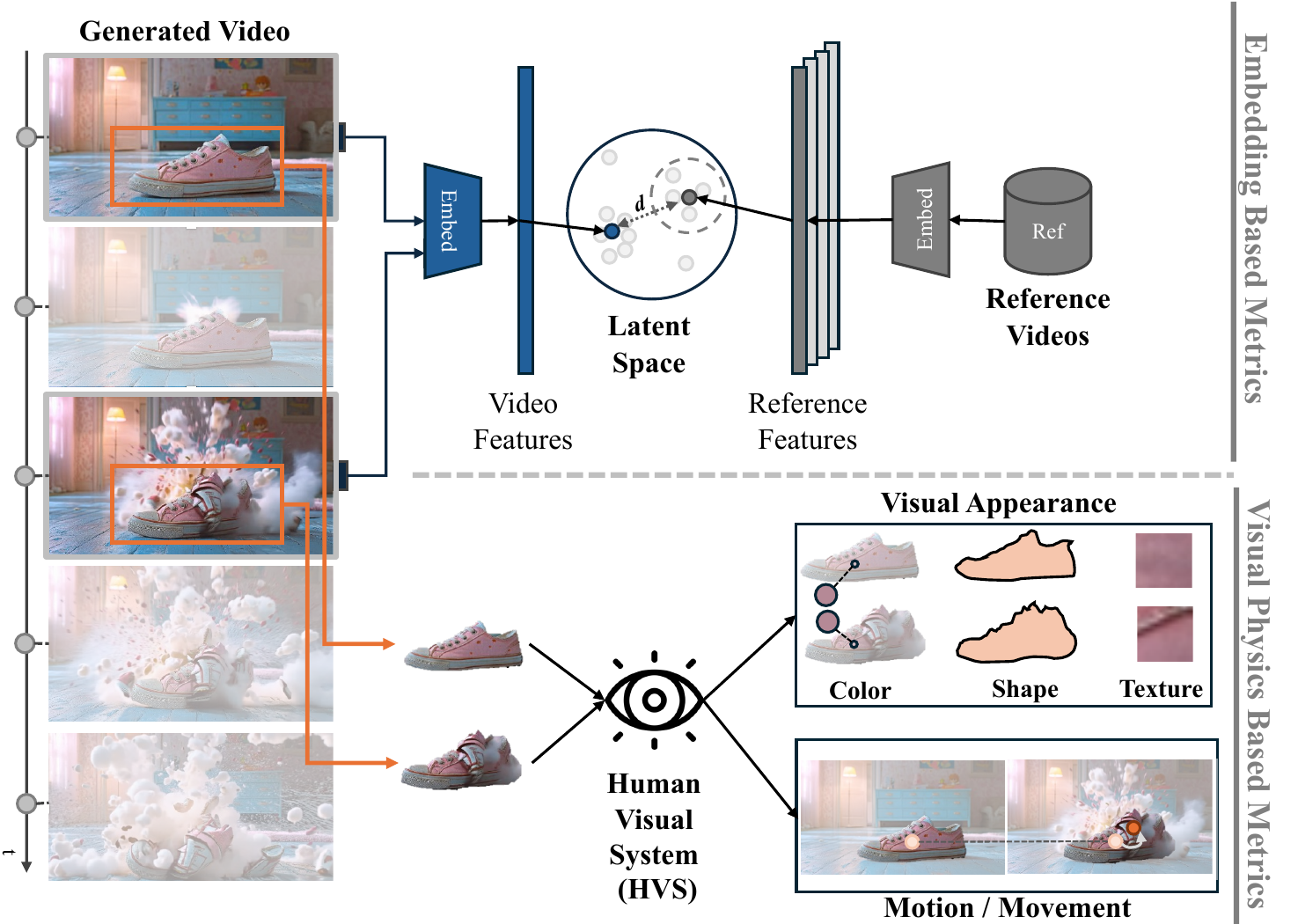} 
    % \fbox{\rule{0pt}{2in} \rule{0.8\linewidth}{0pt}} % Placeholder box
    % \caption{\textbf{Comparison of Embedding-Based and HVS-Inspired Metrics.} The figure contrasts embedding-based metrics, which evaluate video quality in latent space, with HVS-inspired metrics that assess visual appearance (color, shape, texture) and motion coherence, aligning with human visual perception.}
    \caption{\textbf{Comparison of Embedding-Based Metrics and Visual Physics-Based Metrics. }The figure illustrates the evaluation pipelines for generated video quality. The top section represents embedding-based metrics, which rely on extracting features from generated videos and comparing them in a latent space against reference videos. The bottom section introduces a human-visual-system-inspired (HVS) evaluation framework, which analyzes videos based on visual appearance (color, shape, texture) and motion coherence (realistic movement dynamics).}
    \label{fig:metric_overview}
    \vspace{-12pt}
\end{figure}

Recent advancements in video generation, particularly through diffusion models, have dramatically increased the potential for producing highly realistic and dynamic video content {\cite{karras2021alias, ho2022video}}. These models enable sophisticated video synthesis, opening up applications in movie making, virtual reality, autonomous systems, and robotics \cite{qin2024worldsimbenchvideogenerationmodels,videoworldsimulators2024, fu2024exploringinterplayvideogeneration, wu2023unleashing, wang2023drivedreamer, wang2024driving}. However, despite these advancements, evaluating the quality of generated videos remains challenging. Existing metrics, such as Fréchet Video Distance (FVD) {\cite{unterthiner2018towards}}, Inception Score (IS) \cite{salimans2016improved}
and ClipSim {\cite{radford2021learning}}, offer only partial insights into video quality and suffer from significant limitations. For instance, FVD assesses statistical similarity between real and generated videos but lacks interpretability, making its scores difficult to intuitively understand in terms of video quality, while ClipSim captures high-level semantic alignment but can miss finer details, such as texture quality, minor artifacts, or slight color mismatches, which could affect human perception of quality but not be reflected in the score. As video generation models advance in both complexity and realism, these limitations become increasingly problematic. To truly evaluate the quality of generated videos, a comprehensive metric should go beyond latent representations and instead align with standards rooted in human visual perception. Human perception encompasses both visual appearance—covering aspects like color, texture, and shape consistency—and motion coherence, ensuring that movement appears realistic and adheres to natural dynamics. This motivates the development of a new evaluation framework that incorporates both the visual and physical dimensions of video quality, capturing the nuanced standards by which humans assess video realism {\cite{xu2022videogpt}}.

Evaluating the quality of generated videos introduces a set of unique and complex challenges, distinct from the assessment of static images. Unlike images, videos require not only high visual fidelity but also temporal consistency, ensuring coherent visual appearance and with and smooth transitions across frames {\cite{tulyakov2018mocogan}}. Capturing this temporal consistency and coherence is essential, as even minor inconsistencies in visual appearance and movement can break the realism of generated videos. Furthermore, generated videos must adhere to the laws of physics, maintaining geometric consistency within scenes and ensuring that objects and movements behave in a physically plausible manner {\cite{yu2022physics}}. Beyond physical coherence, semantic consistency is crucial—the content and actions depicted in a video must be logically meaningful and align with real-world expectations {\cite{li2022video}}. This involves not only assessing visual quality but also ensuring that the video's narrative or message makes sense. In some cases, particularly when audio is integrated, the evaluation process must also account for multimodal coherence, considering both visual and auditory elements where applicable {\cite{zhou2021cocogan}}. These interconnected challenges—spanning visual, physical, and semantic dimensions—highlight the difficulty of developing a truly comprehensive evaluation metric for generated videos.

To address these challenges, we propose a novel metric designed specifically to evaluate generated video quality, with a focus on both visual and motion consistencies. Grounded in a human visual perspective, this metric offers a more intuitive and comprehensive measure of video quality.  Our contributions are as follows:

\begin{itemize}
    \item We introduce a new video quality evaluation metric that directly targets both visual appearance consistency and motion consistency in generated videos. By moving beyond the limitations of existing metrics like FVD and CLIP score, our approach emphasizes temporal visual appearance and motion realism to provide a more accurate reflection of video quality.

    \item We create a corrupted video dataset which includes five types of corruption: brightness, Gaussian noise, impulse noise, black box, and defocus blur. Each type of corruption is applied at five distinct levels of severity, allowing for a granular analysis of how different types and intensities of distortions impact video quality assessment for our and other metrics.

    \item We conduct extensive experiments on both real and synthetic video datasets, applying our metric to state-of-the-art video generation models. This analysis provides an in-depth comparison of different metrics to measure the quality of the videos, offering insights into their performance on various types of video data. We show that our proposed metric provides a nuanced and more accurate assessment of generated video quality compared to existing approaches. 
\end{itemize}

By addressing these key dimensions—visual appearance and motion consistencies—our work offers a valuable tool to measure the quality of video generation. We expect that this metric will provide a robust foundation for future advancements of video generation, supporting researchers in the pursuit of more realistic and coherent generated video content {\cite{lee2023make}}.

% \begin{figure*}[ht]
%     \centering
%     % Uncomment the following line when you have the actual image file
%     % \includegraphics[width=0.8\textwidth]{path_to_image} 
%     \fbox{\rule{0pt}{2in} \rule{0.8\textwidth}{0pt}} % Placeholder box
%     \caption{An overview of the VAMP (Visual Appearance and Motion Plausibility) metric, illustrating how it evaluates both the visual consistency and motion coherence of generated videos.}
%     \label{fig:metric_overview}
% \end{figure*}

\section{Related Work}
% \textcolor{blue}{\textbf{Need to do more literature review}}

\textbf{Image Generation}
The field of image generation has seen remarkable progress in recent years, primarily driven by advances in deep learning techniques. Generative Adversarial Networks (GANs) \cite{goodfellow2014generative} have been at the forefront of this revolution, offering a wide range of applications \cite{gui2020review, radford2015unsupervised, isola2017image, hu2017toward, yu2017seqgan, wang2022weakly, cao2022learning, duan2017one}. Architectures such as StyleGAN \cite{karras2019style} and BigGAN \cite{brock2018large} have pushed the boundaries of image quality and diversity. More recently, diffusion models \cite{ho2020denoising, zhang2023adding, rombach2022high, bar2023multidiffusion, zhang2023survey} have emerged as a powerful alternative, demonstrating exceptional image quality and flexibility in tasks such as text-to-image generation \cite{ramesh2022hierarchical}\cite{esser2024scalingrectifiedflowtransformers}. These advancements in image generation have laid the groundwork for tackling the more complex task of video generation, providing crucial insights into high-dimensional data synthesis and the importance of perceptual quality in generated content. 

\textbf{Video Generation}
Building upon the success of image generation models, video generation has become an active area of research, presenting unique challenges due to the temporal dimension. Early approaches extended GANs to the video domain, with models like VGAN \cite{vondrick2016generating} and MoCoGAN \cite{tulyakov2018mocogan} attempting to capture both motion and content. Recent work has explored the use of 3D convolutions and attention mechanisms, as seen in Video Transformer Network \cite{li2022video}, to better model long-range dependencies in video sequences. The advent of large language models has also influenced video generation, with text-to-video models like Make-A-Video \cite{lee2023make} demonstrating impressive results in generating videos from textual descriptions. Exploration on using diffusion models led to improvement on text-guided video generation quality \cite{zhou2023magicvideoefficientvideogeneration, ho2022video, ho2022imagen, yang2023diffusion, blattmann2023stable, luo2023videofusion, xing2024survey}.  Despite these advancements, consistent long-term motion and adherence to physical laws remain significant challenges in the field of video generation.

\textbf{Metrics for Video Generation}
As video generation models have evolved, so too has the need for robust evaluation metrics. Traditional image quality metrics like Peak Signal-to-Noise Ratio (PSNR) and Structural Similarity Index (SSIM) have been adapted for video \cite{wang2004video}, but they often fail to capture the complexities of human perception in video quality assessment. The Fréchet Video Distance (FVD) \cite{unterthiner2018towards}, an extension of the Fréchet Inception Distance (FID) used in image generation, has become a popular metric for evaluating the quality and diversity of generated videos. More recently, perceptual metrics leveraging large pre-trained models, such as CLIP score \cite{radford2021learning}, have been applied to video evaluation, offering a more semantic assessment of generated content. Leveraging multi-modal language model and large-scale human feedback to develop evaler demonstrate another promising direction \cite{he2024videoscorebuildingautomaticmetrics}. Some recent emergent metrics take a similar approach as FVD, measuring distances in the latent space with a more detailed focus on specific classes of characteristics \cite{huang2024vbench, liu2024evalcrafter, chivileva2023measuring, li2024sora}. However, they still struggle to fully capture aspects such as temporal consistency, physical plausibility, and long-term coherence in generated videos. Our work aims to address these limitations by introducing a novel metric that specifically targets these crucial aspects of video quality.

% \begin{wrapfigure}{r}{0.4\textwidth} % Reduce the width to allow better text flow
%     \centering
%     % Uncomment the following line when you have the actual image file
%     % \includegraphics[width=0.38\textwidth]{path_to_image} 
%     \fbox{\rule{0pt}{1.5in} \rule{0.38\textwidth}{0pt}} % Placeholder box
%     \caption{Illustration of the point sampling and mask selection process, demonstrating how various sampling methods guide the segmentation in video frames.}
%     \label{fig:point_sampling}
% \end{wrapfigure}

% 1) Video Generation Metrics
% 2) Physical Plausibility and Geometric Consistency in Video Synthesis
% 3) Temporal Coherence in Video Generation
\section{VAMP: \underline{V}isual \underline{A}ppearance and \underline{M}otion \underline{P}lausibility Metric for Video Quality Assessment}

To effectively assess the quality of generated videos, it requires metrics that can capture both visual fidelity and motion coherence, aligning with human perception. We introduce VAMP, a novel metric designed to address both visual appearance and motion consistencies of generated videos. VAMP has three main steps: point sampling, mask selection and VAMP score construction.

\subsection{Point Sampling and Mask Selection}

% The first step of our method employs several point sampling strategies to generate input prompts for the Segment Anything Model 2 (SAM2) \cite{ravi2024sam2}, which is used to produce mask segmentations for video frames:

% \begin{itemize}
%     \item \textbf{Random Sampling:} Points are uniformly distributed across the frame, ensuring broad coverage. The number of points can be adjusted to suit the task.
    
%     \item \textbf{Equally Spaced Sampling:} Points are arranged in a grid pattern. The x-axis point count is configurable, and the y-axis point count is adjusted to maintain the aspect ratio, ensuring even coverage across the frame.
    
%     \item \textbf{SIFT-based Sampling:} The Scale-Invariant Feature Transform (SIFT) \cite{lowe1999object} algorithm is used to detect keypoints in the frame. These feature-rich keypoints are selected as sampling locations.
    
%     \item \textbf{SAM2-based Sampling:} SAM2’s automatic mask generator identifies regions of interest, and the centroids of these regions are selected as sampling points.
% \end{itemize}

% These sampled points act as input prompts for the SAM2 video predictor. The predictor initializes with the first frame, sequentially processes the sampled points, and generates multiple object segmentations. These initial segmentations are then propagated across subsequent frames, enabling the tracking and segmentation of objects throughout the video sequence.

% \textcolor{blue}{\textbf{[Add mask selection part, which is a technical contribution. The mask selection can use the prompt as the prior, $p(masks|prompt)$]}}
To properly track and identify the region of interest in the image to evaluate, we use the above mentioned point sampling methods to identify the important regions to track. We then use SAM2 \cite{ravi2024sam2} to track the movement of those important regions. We mainly experimented with two selection methods: 1) SAM2-based sampling and 2) SIFT-based Sampling, which have trade-off between quality and efficiency. 

One natural way of identifying the regions of interest is utilizing SAM2's auto segmentation ability where SAM2 identify key features in the initial images. We then use the centroids of the key regions as input points to initiate the tracking. The segmentation masks of the regions of interest are then used for downstream calculations. 

An alternate to SAM2 is to use light-weight image features to identify regions of interest. We propose using a combination of SIFT and DBSCAN clustering \cite{ester1996density}. DBSCAN groups packed SIFT features together and drop outliers. It allows us to automatically merge detected points together forming variable numbers of regions instead of pre-defined numbers of regions. As this is a CPU-only process, we can alleviate the compute burden on GPU. We show the results of this method in the supplement material.

% \textcolor{blue}{\textbf{[Design point sampling and mask selection to be an iterative process. If the masks are not good, it can iterate back to point sampling and redo mask selection. When to resample / drop masks? easy cases: masks are gone, masks are converging overlapping, converging to the same region leading to imbalance calculation over a small area.]}}

\subsection{VAMP Score}
\begin{figure*}[ht]
    \centering
    % Uncomment the following line when you have the actual image file
    \includegraphics[width=1\textwidth]{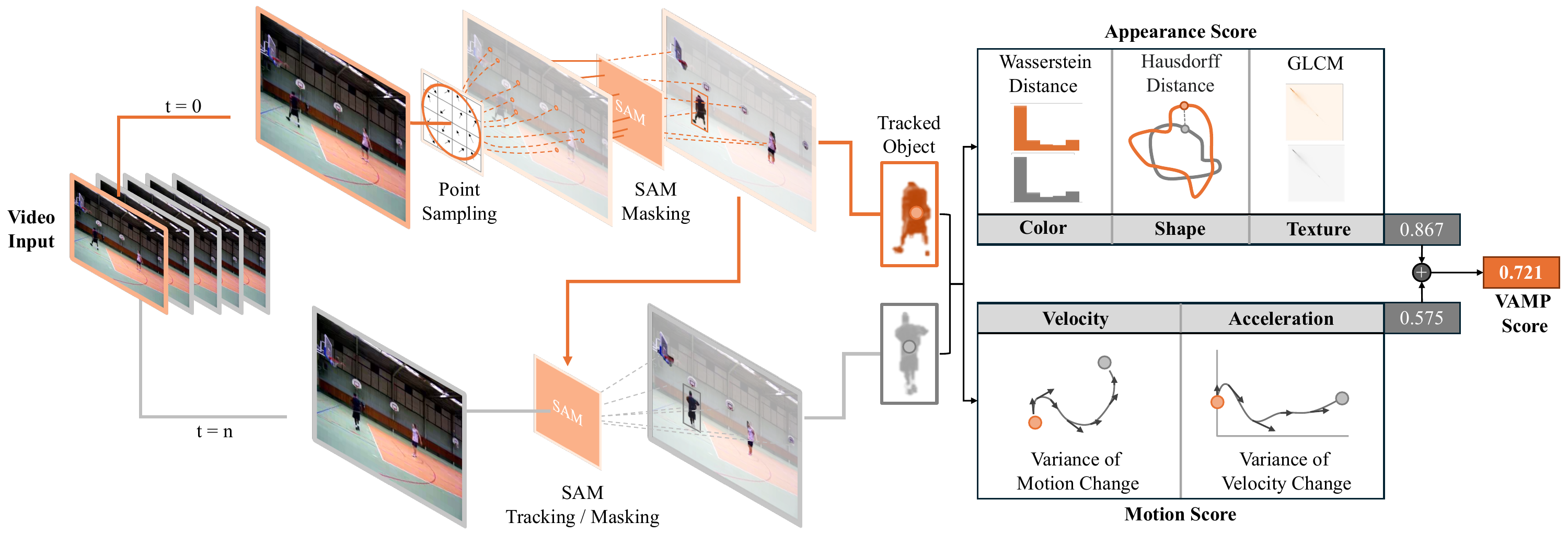} 
    % \fbox{\rule{0pt}{2in} \rule{0.8\textwidth}{0pt}} % Placeholder box
    \caption{\textbf{Pipeline for the VAMP Score Calculation.} This diagram illustrates the process for calculating the VAMP score. Starting with video input, point sampling and SAM masking are applied to identify and track objects across frames. The pipeline computes the Appearance Score, incorporating metrics for color similarity , shape similarity, and texture similarity. Concurrently, the Motion Score is calculated by evaluating velocity and acceleration consistency, capturing the physical plausibility of object movements. These two scores are combined using weighted factors to compute the final VAMP score.}
    \label{fig:metric_overview}
    \vspace{-10pt}
\end{figure*}
Our proposed metric for evaluating generated video quality consists of two main components: an \textit{appearance score} and a \textit{motion score}. Together, these components offer a comprehensive evaluation of both the visual appearance consistency and the motion consistency of generated video sequences. The appearance score captures visual consistency in terms of color, shape, and texture across frames, while the motion score assesses the smoothness of movements of individual objects across frames. 

% \textcolor{blue}{\textbf{Q: How to argue the contribution of sift sampling? Run the formal experiments for the sift and sam2 point sampling, we can compare: 1) similar performance with sift and sam2; 2) faster, 3) less computational cost}}

% \textcolor{brown}{\textbf{[How can we make the metric to be used as a cost to enable gradient descent?]}}

\subsubsection{Appearance Score}
Among studies of the human visual system, color, shape, and texture are three fundamental attributes that play a critical role in visual perception \cite{ge2022contributions}. These attributes form the basis for how humans interpret and recognize objects and scenes in their environment. Color provides information about material properties and lighting conditions, shape conveys structural and spatial information, and texture offers fine-grained details about object surface characteristics. Together, they enable humans to perceive, differentiate, and understand objects and surrounding environments with accuracy, making them essential components for evaluating the visual fidelity of generated videos. 

Therefore, we design an appearance score to measure the visual consistency of objects across frames. The appearance score consists of three sub-scores that assess color similarity, shape similarity, and texture similarity for each detected objects across all frames of the video. These sub-scores work together to ensure that the objects in frames not only appear realistic individually but also maintain visual coherence when viewed in sequence.

\textbf{Color Similarity.} Color similarity is evaluated by comparing the color distribution between consecutive frames. To quantify this, we employ the Earth Mover's Distance (EMD) \cite{rubner2000earth}, which is also known as Wasserstein distance, to measure the dissimilarity between the color distributions between corresponding objects:
\begin{equation}
\text{EMD}(O_{i, t}, O_{i, t+1}) = \min_{\mathbf{f}} \sum_{j,k} f_{j,k} d_{j,k},
\end{equation}
where \(O_{i, t}\) and \(O_{i, t+1}\) are the normalized color histograms of the same corresponding object, $O_i$, across frames at time $t$ and $t+1$, \(f_{j,k}\) represents the flow between bins \(j\) and \(k\), and \(d_{j,k}\) denotes the ground distance between histogram bins. Lower EMD values indicate greater similarity between the color distributions of objects in consecutive frames. EMD is particularly suited for measuring color similarity because it accounts for the spatial arrangement of color distributions, which ensures that subtle shifts in color distributions can be captured. Since there are three channels of the object image, color similarity score is
\begin{equation}
    S_{color}(O_{i, t}, O_{i, t+1}) = 1 - \frac{1}{3}\sum_{c=\{r, g, b\}} \text{EMD}(O_{i, t, c}, O_{i, t+1, c})
\end{equation}
By using EMD, the color similarity score captures the perceptual alignment of colors, ensuring that generated videos maintain realistic and consistent color fidelity across frames.

% We offer two approaches for this comparison: histogram correlation and Earth Mover's Distance (EMD). For histogram comparison, we compute:

% \begin{equation}
%     S_{color} = \text{correl}(H_1, H_2)
% \end{equation}

% where $H_1$ and $H_2$ are the normalized color histograms of two consecutive frames, and $\text{correl}(\cdot)$ represents the correlation between the histograms. Alternatively, we can compute color similarity using EMD:

% \begin{equation}
%     S_{color} = 1 - \frac{1}{3}\sum_{i=1}^3 W(F_{1,i}, F_{2,i})
% \end{equation}

% where $W(\cdot)$ is the Wasserstein distance, and $F_{1,i}$ and $F_{2,i}$ are the flattened and normalized color channels of the two frames.
\textbf{Shape Similarity.} Shape similarity ensures that objects within consecutive frames maintain their structural integrity and geometry. To evaluate shape similarity, we use the Hausdorff distance, a metric for comparing the geometric consistency of object contours. This measure captures both global and fine-grained shape differences, ensuring that the structural properties of objects remain consistent across frames.

The directed Hausdorff distance between two point sets of corresponding object $O_i$ at time $t$, \(O_{i, t}\), and object at time $t+1$, \(O_{i, t+1}\) is defined as:
\begin{equation}
h(O_{i, t}, O_{i, t+1}) = \max_{o_{i,t} \in O_{i,t}} \min_{o_{i, t+1} \in O_{i,t+1}} \|o_{i,t} - o_{i,t+1}\|,
\end{equation}
where \(O_{i,t} = \{o_{i,t,1}, c_{i,t,2}, \dots, c_{i,t,n}\}\) and \(O_{i,t+1} = \{o_{i,t+1,1}, o_{i,t+1,2}, \dots, o_{i,t+1,m}\}\) are the sets of points representing the contours of the objects in two consecutive frames, and \(\|\cdot\|\) denotes the Euclidean distance between points.

The final shape similarity score is computed using the symmetric Hausdorff distance, which considers the maximum of the directed distances in both directions:
\begin{equation}
S_{\text{Hausdorff}} = \frac{1}{1 + \max(h(O_{i, t}, O_{i, t+1}), h(O_{i, t+1}, O_{i, t}))},
\end{equation}
and the shape similarity score can be written as:
\begin{equation}
S_{shape}(O_{i,t}, O_{i,t+1}) = \frac{1} {1 + S_{\text{Hausdorff}}(O_{i,t}, O_{i,t+1})}. 
\end{equation}
By using the Hausdorff distance, the shape similarity score effectively captures the geometric alignment of object contours across frames, ensuring structural consistency throughout the video.

\textbf{Texture Similarity.} Texture similarity assesses the fine-grained surface details of objects across consecutive frames, ensuring that surface patterns remain consistent throughout the video. To evaluate texture similarity, we resize the images of corresponding objects to a common target size and convert them to grayscale for uniform processing. We then compute texture features using the Gray Level Co-occurrence Matrix (GLCM), which captures spatial relationships between pixel intensities, a key indicator of texture.

The GLCM is calculated for multiple orientations (\(0\), \(\pi/4\), \(\pi/2\), and \(3\pi/4\)) and a distance of 1 pixel, with intensity levels quantized to 256 for consistency. From the GLCM, we extract five statistical properties—contrast, dissimilarity, homogeneity, energy, and correlation—that comprehensively describe texture. These properties are flattened into feature vectors for each object in consecutive frames. 

The similarity between the texture features of two frames is then computed using cosine similarity:
\begin{equation}
S_{\text{texture}}(O_{i,t}, O_{i,t+1}) = \frac{\mathbf{f}_{O_{i,t}} \cdot \mathbf{f}_{O_{i,t+1}}}{\|\mathbf{f}_{O_{i,t}}\| \|\mathbf{f}_{O_{i,t+1}}\|},
\end{equation}
where \(\mathbf{f}_{O_{i,t}}\) and \(\mathbf{f}_{O_{i,t+1}}\) are the feature vectors derived from the GLCM properties of the corresponding objects at time $t$ and $t+1$. Higher cosine similarity values indicate greater alignment of texture details between frames. This approach ensures that texture consistency can be accurately captured.

\textbf{Appearance Score.} The overall appearance score evaluates the visual fidelity and consistency of objects across frames by integrating three key components: color similarity, shape similarity, and texture similarity. These components collectively ensure that the generated video maintains coherence in color distribution, geometric structure, and fine-grained surface details throughout the sequence.

The final appearance score is computed as the weighted average of the individual similarity scores:
\begin{equation}
S_{\text{appearance}} = w_{\text{c}} S_{\text{color}} + w_{\text{s}} S_{\text{shape}} + w_{\text{t}} S_{\text{texture}},
\end{equation}
where \(S_{\text{color}}, S_{\text{shape}}, \text{and } S_{\text{texture}}\) are the respective similarity scores for color, shape, and texture, and \(w_{\text{c}}, w_{\text{s}}, \text{and } w_{\text{t}}\) are their corresponding weights, normalized such that:
\begin{equation}
w_{\text{color}} + w_{\text{shape}} + w_{\text{texture}} = 1.
\end{equation}
These weights can be tuned based on the importance of each component in the specific application. By combining these measures, the overall appearance score provides a comprehensive evaluation of the visual quality of generated videos, aligning with human perception of visual coherence.

% \textbf{Texture Similarity.} Texture similarity assesses the fine-grained surface details of objects across consecutive frames. We provide two methods for texture comparison: Gray Level Co-occurrence Matrix (GLCM) features or Local Binary Patterns (LBP). For GLCM-based similarity, we compute:

% \begin{equation}
%     S_{texture} = \frac{\mathbf{f}_1 \cdot \mathbf{f}_2}{\|\mathbf{f}_1\| \|\mathbf{f}_2\|}
% \end{equation}

% where $\mathbf{f}_1$ and $\mathbf{f}_2$ represent vectors of GLCM properties (contrast, dissimilarity, homogeneity, energy, correlation) for consecutive frames. Alternatively, using LBP:

% \begin{equation}
%     S_{texture} = \frac{1}{1 + \sum_{i=1}^{256} \frac{(H_{1,i} - H_{2,i})^2}{H_{1,i} + H_{2,i} + \epsilon}}
% \end{equation}

% where $H_{1,i}$ and $H_{2,i}$ are the normalized LBP histograms of two consecutive frames. By combining these three sub-scores—color, shape, and texture—the appearance score offers a well-rounded assessment of visual fidelity.

% \textcolor{blue}{\textbf{[Normalization based on real world values/distributions]}}

\subsubsection{Motion Score}

The motion score evaluates the physical plausibility of object movements in the generated video. It incorporates several components designed to capture the realism of motion, including velocity consistency and acceleration consistency. These components ensure that movements adhere to natural laws and that the generated video maintains temporal coherence.

\textbf{Velocity Consistency.} Velocity consistency evaluates the stability of object motion by analyzing the trajectories of object centroids across frames. For each object, the velocities are calculated as the differences between the centroids of consecutive frames. The average distance between these centroids is measured using:
\begin{equation}
v_t = \|\mathbf{c}_{t+1} - \mathbf{c}_t\|,
\end{equation}
where \(\mathbf{c}_t\) and \(\mathbf{c}_{t+1}\) are the centroids of the object in frames \(t\) and \(t+1\). To quantify the consistency of velocity, we compute the standard deviation of the velocity values across all frames and normalize it by the mean velocity. The velocity consistency score is then defined as:
\begin{equation}
S_{\text{vel}} = \exp\left(-\frac{\text{std}(\mathbf{v})}{\text{mean}(\mathbf{v})}\right),
\end{equation}
Lower velocity variance indicates smoother and more consistent motion.

\textbf{Acceleration Consistency.} Acceleration consistency evaluates the smoothness of changes in velocity across frames. For each object, accelerations are computed as the differences between velocities in consecutive frames:
\begin{equation}
a_t = v_{t+1} - v_t,
\end{equation}
where \(v_t\) and \(v_{t+1}\) represent the velocities between frames \(t\) and \(t+1\). The variance of accelerations is then calculated, as lower acceleration variance indicates smoother motion transitions. The acceleration consistency score is defined as:
\begin{equation}
S_{\text{acc}} = \exp(-\text{Var}(\mathbf{a})),
\end{equation}
where \(\mathbf{a}\) is the vector of accelerations across all frames for a given object.

\textbf{Motion Score.} The final motion score for each object is computed as the weighted combination of the individual components:
\begin{equation}
S_{\text{motion}} = w_{\text{vel}} S_{\text{vel}} + w_{\text{acc}} S_{\text{acc}},
\end{equation}
where \(w_{\text{vel}}, w_{\text{acc}}, \text{and } w_{\text{depth}}\) are the weights for velocity, acceleration, and depth consistency, respectively, satisfying:
\begin{equation}
w_{\text{vel}} + w_{\text{acc}} = 1.
\end{equation}
By integrating these components, the motion score provides a comprehensive assessment of the physical plausibility and temporal coherence of object movements in generated videos.

\begin{figure*}[!ht]
    \centering
    \includegraphics[width=0.95\textwidth]{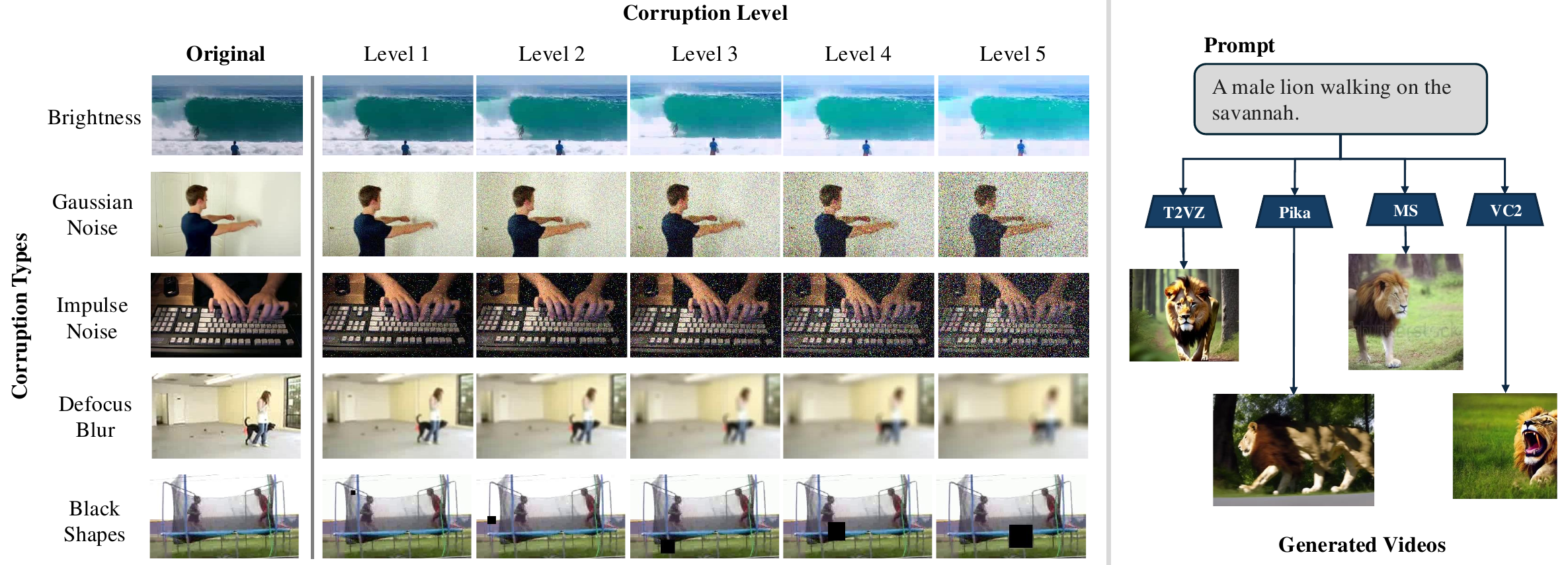} % Adjust the width as needed
    \caption{Evaluation of video quality using the VAMP metric across different levels of corruptions and generative models. This figure shows the impact of various corruption types on video quality as assessed by the VAMP metric. (Left) Examples of corrupted videos showing the effect of Brightness, Gaussian Noise, Impulse Noise, Defocus Blur, and Black Shapes at different severity levels. (Right) Generated video outputs from state-of-the-art video generation models (T2VZ, Pika, MS, VC2) based on the given prompt.}
    \label{fig:results}
    \vspace{-11pt}
\end{figure*}

\subsubsection{VAMP Score}

By combining the visual appearance and motion scores, our proposed metric provides a comprehensive evaluation of video generation quality, ensuring that both visual fidelity and physical plausibility are considered. This holistic approach allows for more accurate and intuitive assessments of generated video content. 

The VAMP score of a video is computed as a weighted sum of the appearance score (\(S_{\text{appearance}}\)) and the motion score (\(S_{\text{motion}}\)) across all the frames and objects:
\begin{equation}
S_{\text{VAMP}} = \frac{1}{mn}\sum_{t=0}^n\sum_{i=0}^m\alpha S_{\text{appearance}, i, (t, t+1)} + \beta S_{\text{motion}, i, (t, t+1)}
\end{equation}
where $m$ is the total number of objects, $n$ is total number of frames, \(\alpha\) and \(\beta\) are the weights assigned to the appearance and motion scores, respectively, and satisfy the normalization condition:
\begin{equation}
\alpha + \beta = 1.
\end{equation}
These weights allow the metric to be adjusted based on the specific requirements of the application. For instance, applications where visual fidelity is more critical may assign a higher value to \(\alpha\), while scenarios requiring precise motion realism may prioritize \(\beta\).

\textbf{Interpretability.} A higher \(S_{\text{VAMP}}\) score reflects better alignment with human perception of video quality, encompassing both visual appearance and object movement aspects. This makes \(S_{\text{VAMP}}\) an intuitive and practical metric for evaluating the quality of generated video content across diverse domains. By combining these complementary dimensions, the VAMP score bridges the gap between traditional evaluation metrics and the nuanced expectations of human observers, offering a robust foundation for benchmarking video generation models.

% \subsubsection{VAMP Score}

% By combining the visual appearance and motion scores, our proposed metric provides a comprehensive evaluation of video generation quality, ensuring that both visual fidelity and physical plausibility are considered. This holistic approach allows for more accurate and meaningful assessments of generated video content.

% \begin{equation}
%     S_{VAMP} = \alpha S_{appearance} + \beta S_{motion} 
% \end{equation}

% \textcolor{blue}{\textbf{[IMPORTANT!!!: Add the final metric construction]}}

\section{Experiments}
\label{experiments}

\begin{table*}[!t]
\centering
{\scriptsize
\begin{subtable}[t]{\textwidth}
    \centering
    \begin{tabular}{lcccccc|cccccc}
    \toprule
    \multirow{2}{*}{\textbf{Method}} & \multicolumn{6}{c|}{\textbf{Brightness}} & \multicolumn{6}{c}{\textbf{Gaussian Noise}} \\ 
    \cmidrule(lr){2-7} \cmidrule(lr){8-13} 
   &0 & 1 & 2 & 3 & 4 & 5 
   &0 & 1 & 2 & 3 & 4 & 5 \\ 
    \midrule
    ClipSim ↑ &0.969 & 0.968 & 0.967 & 0.966 & 0.965 & 0.965 &0.969 & 0.948 & 0.944 & 0.945 & 0.945 & 0.944  \\
    IS ↑      &1.470& 1.494 & 1.506 & 1.523 & 1.533 & 1.537 &1.470& 1.555 & 1.628 & 1.721 & 1.829 & 1.816 \\
    FVD ($\times 10^3$) ↓ &0 & 0.239 & 0.657 & 1.121 & 1.721 & 2.429 &0 & 1.759 & 2.974 & 4.295 & 5.395 & 6.314 \\
    \midrule
    VAMP-A ↑  &0.726& 0.729 & 0.713 & 0.727 & 0.721 & 0.712 &0.726& 0.694 & 0.661 & 0.663 & 0.562 & 0.541 \\
    VAMP-M ↑ &0.667 & 0.645 & 0.632 & 0.624 & 0.613 & 0.607 &0.667& 0.604 & 0.565 & 0.519 & 0.439 & 0.388 \\
    \midrule
    VAMP ↑   &0.685 & 0.670 & 0.656 & 0.655 & 0.646 & 0.638 &0.685 & 0.631 & 0.594 & 0.562 & 0.476 & 0.434 \\
    \bottomrule
    \end{tabular}
    \label{tab:main_corruption}
\end{subtable}

\vspace{1em} % Add space between the two subtables

\begin{subtable}[t]{\textwidth}
    \centering
    \begin{tabular}{cccccc|cccccc|cccccc}
    \toprule
    \multicolumn{6}{c|}{\textbf{Impulse Noise}} & \multicolumn{6}{c|}{\textbf{Defocus Blur}} & \multicolumn{6}{c}{\textbf{Black Shapes}} \\ 
    \cmidrule(lr){1-6} \cmidrule(lr){7-12} \cmidrule(lr){13-18}
    0 & 1 & 2 & 3 & 4 & 5 
    & 0 & 1 & 2 & 3 & 4 & 5 
    & 0 & 1 & 2 & 3 & 4 & 5 \\ 
    \midrule
    0.969 & 0.948 & 0.946 & 0.946 & 0.943 & 0.943 
    & 0.969 & 0.970 & 0.970 & 0.972 & 0.973 & 0.975 
    & 0.969 & 0.962 & 0.955 & 0.951 & 0.948 & 0.944 \\ 
    1.470 & 1.679 & 1.728 & 1.775 & 1.876 & 1.846 
    & 1.470 & 1.418 & 1.421 & 1.419 & 1.395 & 1.367 
    & 1.470 & 1.496 & 1.509 & 1.539 & 1.561 & 1.596 \\ 
    0 & 3.515 & 4.350 & 4.885 & 5.789 & 6.431 
    & 0 & 1.244 & 1.949 & 3.355 & 4.575 & 5.290 
    & 0 & 2.362 & 1.722 & 1.578 & 1.866 & 2.230 \\ 
    \midrule
    0.726 & 0.726 & 0.711 & 0.699 & 0.660 & 0.582 
    & 0.726 & 0.721 & 0.698 & 0.580 & 0.372 & 0.220 
    & 0.726 & 0.683 & 0.681 & 0.659 & 0.647 & 0.642 \\ 
    0.667 & 0.621 & 0.594 & 0.566 & 0.501 & 0.404 
    & 0.667 & 0.640 & 0.605 & 0.477 & 0.281 & 0.154 
    & 0.667 & 0.636 & 0.573 & 0.533 & 0.523 & 0.519 \\ 
    \midrule
    0.685 & 0.653 & 0.629 & 0.606 & 0.549 & 0.458 
    & 0.685 & 0.664 & 0.633 & 0.508 & 0.308 & 0.173 
    & 0.685 & 0.650 & 0.606 & 0.571 & 0.560 & 0.556 \\
    \bottomrule
    \end{tabular}

    \label{tab:brightness_blackshapes}
\end{subtable}
}

\vspace{-5pt}
\caption{Performance of Metrics Across Corruption Types and Levels. 
This table shows the evaluation of video quality metrics under different corruption types (Brightness, Gaussian Noise, Impulse Noise, Defocus Blur, and Black Shapes) applied at levels 1–5. Metrics include ClipSim, IS, FVD ($\times 10^3$), VAMP-A (appearance-only), VAMP-M (motion-only), and VAMP.}
\label{tab:corruption_results}
\vspace{-12pt}
\end{table*}

We evaluate the robustness and effectiveness of our proposed metric in two scenarios: corrupted video evaluation and generated video evaluation. To benchmark its performance, we compare it against established metrics such as Fréchet Video Distance (FVD), Inception Score (IS), and ClipSim. The following experiments demonstrate how our metric captures both visual fidelity and temporal consistency in videos and outperforms baseline methods.

\subsection{Corrupted Video Evaluation}
% \textcolor{blue}{\textbf{[Make the corrupted videos as a dataset benchmark releases with the paper (contribution)]}}

In this experiment, we aim to evaluate the sensitivity of our proposed metric by injecting various types of noise and corruption into real videos. 

\textbf{Dataset.} Given the wide range of possible video scenes and objects, it is impractical to evaluate all types of video content. In this study, we focus on videos involving \textit{humans with actions}, which are among the most common types of objects generated by video generation models. We use the UCF101 dataset \cite{soomro2012ucf101}, which contains a diverse collection of human actions across 101 categories within different scenes, making it ideal for evaluating videos involving human appearance and motion.

\textbf{Corruption Types.} In this experiments, we apply various types of noise to videos from each dataset, including \textit{brightness}, \textit{gaussian noise},\textit{impulse noise}, \textit{defocus blur}, and \textit{black shapes}. Our goal is to measure how well the metric captures the impact of these corruptions across different objects and scene types. The correlation between the noise severity and the metric score will reveal the robustness and effectiveness of the metric in detecting quality degradation in various real-world video contexts. We apply a variety of common image corruption techniques to real videos, simulating different types of noise and distortions. Below is a brief description of each type of corruption used in the experiments :
% \textcolor{blue}{\textbf{[corruption descriptions can be added to the appendix]}}:

\begin{itemize}

    \item \textit{Brightness:} Modifies the overall brightness of the video frames, simulating changes in lighting conditions.
    
    \item \textit{Gaussian Noise:} Adds Gaussian noise to the video frames, simulating random variations in brightness or color. This type of noise is often used to simulate sensor noise.
    
    % \item \textbf{Shot Noise:} Simulates photon counting errors, typically more prominent in dark areas. This type of noise is more likely to affect low-light portions of a video.

    \item \textit{Impulse Noise (Salt-and-Pepper Noise):} Adds random white and black pixels to the video, simulating the effect of faulty pixel readings in a sensor or transmission errors.

    \item \textit{Defocus Blur:} Simulates the effect of the video being out of focus, causing a uniform blur across the frames.

    % \item \textbf{Motion Blur:} Applies a directional blur to simulate the effect of camera or object movement during exposure.

    % \item \textbf{Zoom Blur:} Introduces a radial blur, simulating the effect of quickly zooming in or out during video capture.

    % \item \textbf{Contrast:} Changes the contrast between light and dark areas of the video, affecting how much detail is visible.

    % \item \textbf{Pixelation:} Reduces the resolution of the video frames by increasing the size of individual pixels, giving the video a blocky appearance.

    \item \textit{Black Shapes:} Adds random black boxes to the video frames, simulating occlusions, artifacts, or sensor malfunctions.
\end{itemize}

% \textcolor{blue}{\textbf{[Add (1) Mask Color Corruption and (2) Shape Distortion Corruption]}}

\textbf{Experiment Setup.} In this experiment, we applied noise to a set of real videos at varying levels (1-5) to simulate different degrees of corruption. Each type of corruption was systematically injected into the original videos, altering the visual quality to different extents. After introducing these distortions, we used our proposed metric to evaluate the resulting videos, generating both an appearance score and a motion score for each, which are used to compute VAMP score. Then, we compare the score calculated from our metric with the baseline methods to show the effectiveness on reflecting the correlations between corruption levels and scores. 

% The change in the overall metric score was monitored as the severity of the corruption increased, allowing us to determine the sensitivity of the metric to the different levels of degradation. This approach enabled us to assess how effectively the metric captures the decline in video quality due to the applied noise. Finally, we computed the correlation between the metric scores and the severity of the corruptions, providing quantitative insight into how well the metric reflects the increasing noise levels in the videos.

\subsection{Generated Video Evaluation}

In addition to evaluating our metric on corrupted real videos, we further assess its effectiveness by applying it to videos generated by current state-of-the-art video generation models. 

% \textbf{Prompt Benchmark.} To ensure consistency and diversity in the generated content, we design a comprehensive set of prompts that focus on the most commonly generated objects in video synthesis: humans with actions, animals, and vehicles. This prompt benchmark includes well-crafted prompts that specify scenes involving human activities, dynamic animal behavior, and various vehicles in motion. The goal is to capture a wide range of motions and object interactions that can challenge the generative capabilities of the models.

% \textcolor{blue}{\textbf{[Create a set of prompts for video generation]}}
\textbf{Experiment Setup.} For this evaluation, we sampled 3,000 prompts from the VidProM dataset \cite{wang2024vidprommillionscalerealpromptgallery}, which includes a diverse range of video prompts and scenarios. Four video generation models were used to produce these videos: \textit{Pika} \cite{pikalabs2024}, \textit{VideoCraft2} \cite{chen2024videocrafter2}
, \textit{Text2Video-Zero} \cite{khachatryan2023text2videozerotexttoimagediffusionmodels}
, and \textit{ModelScope} \cite{wang2023modelscopetexttovideotechnicalreport}
. The sampled prompts were used by the video generation models to generate corresponding videos. With the generated videos, we evaluate them using our proposed metric, calculating both the appearance score and motion score for each video. The evaluation provides insights into how well the generated videos maintain visual fidelity and motion coherence across different models and scenarios.

\section{Results and Analysis}
\label{results}

\begin{figure*}[!ht]
    \centering
    \includegraphics[width=0.85\textwidth]{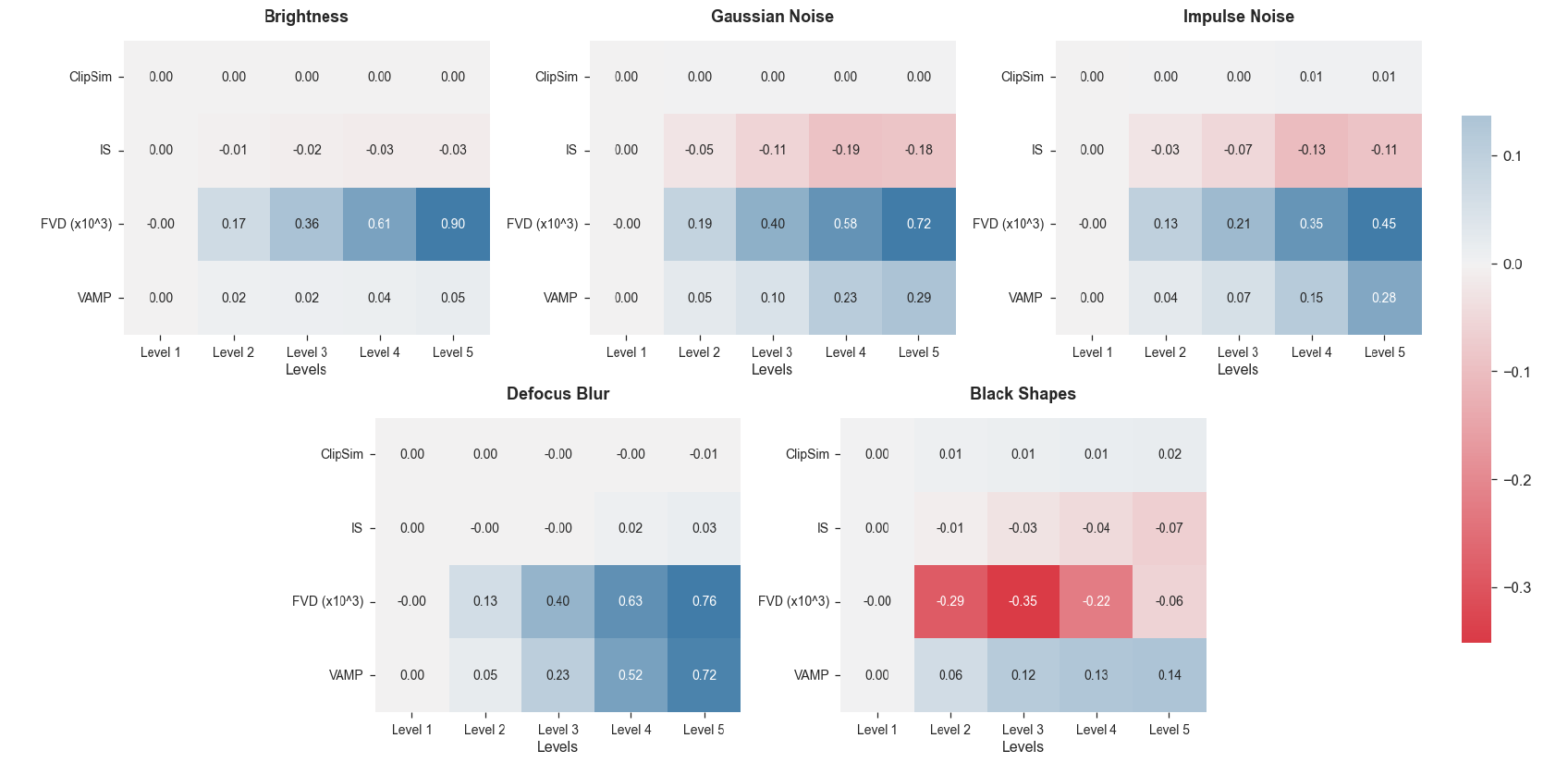}
    \caption{Normalized changes in metric values across corruption types and levels. Each heatmap represents the impact of a specific corruption type on four evaluation metrics (ClipSim, IS, FVD, and VAMP). Blue indicates a positive correlation, while red indicates a negative correlation.}
    \label{fig:normalized_heatmaps}
\vspace{-12pt}
\end{figure*}

\textbf{Results.} The evaluation results on corrupted videos are summarized in Table \ref{tab:corruption_results}, which presents the performance of video quality metrics under different corruption types (Brightness, Gaussian Noise, Impulse Noise, Defocus Blur, and Black Shapes) applied at varying levels (1–5). The table includes scores for ClipSim, IS, FVD, VAMP-A (appearance-only), VAMP-M (motion-only), and VAMP. Table \ref{tab:generative_model_scores_narrow} reports the scores of generative models (Text2Video-Zero, ModelScope, VideoCrafter2, and Pika) across appearance metrics (Color, Shape, Texture), motion score, and the overall VAMP score. Figure \ref{fig:normalized_heatmaps} illustrates the normalized changes in metric values for each corruption type and level relative to the original video, showing the effect of corruption on metrics such as ClipSim, IS, FVD, and VAMP. Based on these results, we highlight some key findings in the following paragraphs regarding the sensitivity of the metrics to different corruption types and the generative models' performance across various metrics.

% \textbf{VAMP score can achieve similar performance as FVD.} The analysis of Table \ref{tab:corruption_results} and Figure \ref{fig:corruption_heatmaps} highlights that the VAMP score performs comparably to FVD across various corruption types and levels. Both metrics consistently reflect increasing degradation, with FVD showing sharper increases, particularly for Gaussian Noise and Impulse Noise. Figure \ref{fig:corruption_heatmaps} indicates that VAMP captures quality changes with balanced sensitivity, especially for challenging corruptions like Black Shapes, where its response is more stable. This demonstrates that VAMP provides a robust and interpretable alternative to FVD, effectively balancing motion and appearance evaluations in video quality assessment.

\textbf{VAMP captures both visual appearance and motion consistencies of videos.} The results presented in Table \ref{tab:corruption_results} and Table \ref{tab:generative_model_scores_narrow} demonstrate that the VAMP metric effectively evaluates both appearance and motion aspects of video quality. For appearance, VAMP-A specifically quantifies attributes such as color, shape, and texture, as evidenced by its sensitivity to degradation levels in Table \ref{tab:corruption_results}. For motion consistency, VAMP-M highlights temporal coherence across frames, maintaining robust scores despite challenging corruptions like Impulse Noise and Defocus Blur. Furthermore, Figure \ref{fig:normalized_heatmaps} illustrates that VAMP balances responses to visual and motion degradations, showing incremental changes across corruption levels that align with human perceptual degradation. These results support VAMP's capability to holistically assess video quality by incorporating both spatial and temporal dimensions.

\textbf{VAMP can be a good reference-free alternative to FVD.} The analysis of Table \ref{tab:corruption_results} and Figure \ref{fig:normalized_heatmaps} demonstrates that the VAMP score performs comparably to FVD across diverse corruption types and levels. Both metrics reliably capture increasing degradation in video quality as corruption levels intensify, with FVD generally exhibiting sharper increases, particularly under Gaussian Noise and Impulse Noise. However, a key advantage of VAMP is its independence from reference videos, unlike FVD, which requires access to ground-truth data for effective evaluation. This feature makes VAMP more versatile and applicable in real-world scenarios where reference videos may not always be available or practical to use. Figure \ref{fig:normalized_heatmaps} further highlights importance of VAMP's evaluation on visual appearance not in latent space. This is particularly evident in corruption types such as Black Shapes, where VAMP exhibits greater stability and consistency in its response compared to FVD, which shows negative correlations to corruption levels. Overall, the VAMP metric emerges as a robust and interpretable alternative to FVD, particularly suited for use cases where reference videos are not accessible. 

\textbf{ClipSim and IS may not be reliable metrics for evaluating generated video quality in certain cases.} As shown in Table \ref{tab:corruption_results}, IS exhibits only minor changes across different corruption levels for Brightness and Gaussian Noise, suggesting it is less sensitive to perceptual degradations. Similarly, ClipSim remains largely unchanged across all tested corruption types, as seen in Figure \ref{fig:normalized_heatmaps}, with negligible variations even for severe distortions like Impulse Noise and Defocus Blur. This insensitivity implies that neither IS nor ClipSim adequately reflects the impact of degradations on visual quality or motion consistency. For example, while Black Shapes significantly affect the motion and appearance of videos, ClipSim fails to capture these changes effectively, maintaining nearly constant values. These limitations highlight that IS and ClipSim might not fully represent the quality of generated videos, particularly in scenarios where both visual appearance and temporal consistency are critical for evaluation.

\begin{table}[ht]
\centering

{\scriptsize
\begin{tabular}{>{\centering\arraybackslash}m{2.5cm}|ccc|c|c}
\toprule
\textbf{Model} & \textbf{Color} & \textbf{Shape} & \textbf{Texture} & \textbf{Motion} & \textbf{VAMP} \\ 
\midrule
Text2Video-Zero & -     & -     & -     & 0.448 & 0.314 \\
ModelScope      & 0.909 & 0.414 & 0.956 & 0.631 & 0.650 \\
VideoCrafter2   & 0.957 & 0.448 & 0.975 & 0.683 & 0.696 \\
Pika            & \textbf{0.993} & \textbf{0.477} & \textbf{0.997} & \textbf{0.788} & \textbf{0.778} \\
\bottomrule
\end{tabular}
}
\vspace{-7pt}
\caption{Quantitative evaluation of generative video models across appearance and motion metrics. The table presents the scores for four video generation models on appearance metrics (Color, Shape, and Texture), Motion Score, and the overall VAMP Score.}
\label{tab:generative_model_scores_narrow}
\vspace{-15pt}
\end{table}

\section{Conclusion}
\label{conclusion}
In this paper, we introduced VAMP, a novel metric for evaluating both visual fidelity and motion coherence in generated videos. By assessing visual appearance consistency and the motion consistency, VAMP provides a more comprehensive evaluation than traditional metrics like IS, and ClipSim, and achieved compatible results as FVD without the need of reference videos. Our experiments show that VAMP effectively captures video quality degradation and aligns well with human evaluations on visual appearance and motion consistency.

\textbf{Limitations and Future Work.} VAMP involves extensive parameter tuning which can be time-consuming and may not generalize well. Its computational complexity presents challenges due to resource-intensive object segmentation and tracking from SAM. Future work can explore adaptive tuning mechanisms and optimize the computational pipeline to enhance efficiency and scalability.

{
    \small
    \bibliographystyle{ieeenat_fullname}
    \bibliography{main}

\begin{thebibliography}{58}
\providecommand{\natexlab}[1]{#1}
\providecommand{\url}[1]{\texttt{#1}}
\expandafter\ifx\csname urlstyle\endcsname\relax
  \providecommand{\doi}[1]{doi: #1}\else
  \providecommand{\doi}{doi: \begingroup \urlstyle{rm}\Url}\fi

\bibitem[Bar-Tal et~al.(2023)Bar-Tal, Yariv, Lipman, and Dekel]{bar2023multidiffusion}
Omer Bar-Tal, Lior Yariv, Yaron Lipman, and Tali Dekel.
\newblock Multidiffusion: Fusing diffusion paths for controlled image generation.
\newblock 2023.

\bibitem[Blattmann et~al.(2023)Blattmann, Dockhorn, Kulal, Mendelevitch, Kilian, Lorenz, Levi, English, Voleti, Letts, et~al.]{blattmann2023stable}
Andreas Blattmann, Tim Dockhorn, Sumith Kulal, Daniel Mendelevitch, Maciej Kilian, Dominik Lorenz, Yam Levi, Zion English, Vikram Voleti, Adam Letts, et~al.
\newblock Stable video diffusion: Scaling latent video diffusion models to large datasets.
\newblock \emph{arXiv preprint arXiv:2311.15127}, 2023.

\bibitem[Brock et~al.(2018)Brock, Donahue, and Simonyan]{brock2018large}
Andrew Brock, Jeff Donahue, and Karen Simonyan.
\newblock Large scale gan training for high fidelity natural image synthesis.
\newblock \emph{arXiv preprint arXiv:1809.11096}, 2018.

\bibitem[Brooks et~al.(2024)Brooks, Peebles, Holmes, DePue, Guo, Jing, Schnurr, Taylor, Luhman, Luhman, Ng, Wang, and Ramesh]{videoworldsimulators2024}
Tim Brooks, Bill Peebles, Connor Holmes, Will DePue, Yufei Guo, Li Jing, David Schnurr, Joe Taylor, Troy Luhman, Eric Luhman, Clarence Ng, Ricky Wang, and Aditya Ramesh.
\newblock Video generation models as world simulators.
\newblock 2024.

\bibitem[Cao et~al.(2022)Cao, Wang, and Sadigh]{cao2022learning}
Zhangjie Cao, Zihan Wang, and Dorsa Sadigh.
\newblock Learning from imperfect demonstrations via adversarial confidence transfer.
\newblock In \emph{2022 International Conference on Robotics and Automation (ICRA)}, pages 441--447. IEEE, 2022.

\bibitem[Chen et~al.(2024)Chen, Zhang, Cun, Xia, Wang, Weng, and Shan]{chen2024videocrafter2}
Haoxin Chen, Yong Zhang, Xiaodong Cun, Menghan Xia, Xintao Wang, Chao Weng, and Ying Shan.
\newblock Videocrafter2: Overcoming data limitations for high-quality video diffusion models, 2024.

\bibitem[Chivileva et~al.(2023)Chivileva, Lynch, Ward, and Smeaton]{chivileva2023measuring}
Iya Chivileva, Philip Lynch, Tomas~E Ward, and Alan~F Smeaton.
\newblock Measuring the quality of text-to-video model outputs: Metrics and dataset.
\newblock \emph{arXiv preprint arXiv:2309.08009}, 2023.

\bibitem[Duan et~al.(2017)Duan, Andrychowicz, Stadie, Ho, Schneider, Sutskever, Abbeel, and Zaremba]{duan2017one}
Yan Duan, Marcin Andrychowicz, Bradly Stadie, Jonathan Ho, Jonas Schneider, Ilya Sutskever, Pieter Abbeel, and Wojciech Zaremba.
\newblock One-shot imitation learning.
\newblock In \emph{Advances in Neural Information Processing Systems (NeurIPS)}, pages 1087--1098, 2017.

\bibitem[Esser et~al.(2024)Esser, Kulal, Blattmann, Entezari, Müller, Saini, Levi, Lorenz, Sauer, Boesel, Podell, Dockhorn, English, Lacey, Goodwin, Marek, and Rombach]{esser2024scalingrectifiedflowtransformers}
Patrick Esser, Sumith Kulal, Andreas Blattmann, Rahim Entezari, Jonas Müller, Harry Saini, Yam Levi, Dominik Lorenz, Axel Sauer, Frederic Boesel, Dustin Podell, Tim Dockhorn, Zion English, Kyle Lacey, Alex Goodwin, Yannik Marek, and Robin Rombach.
\newblock Scaling rectified flow transformers for high-resolution image synthesis, 2024.

\bibitem[Ester et~al.(1996)Ester, Kriegel, Sander, Xu, et~al.]{ester1996density}
Martin Ester, Hans-Peter Kriegel, Jorg Sander, Xiaowei Xu, et~al.
\newblock A density-based algorithm for discovering clusters in large spatial databases with noise.
\newblock In \emph{kdd}, pages 226--231, 1996.

\bibitem[Fu et~al.(2024)Fu, Zhou, Zhou, Yang, Gao, Li, Wu, and Shao]{fu2024exploringinterplayvideogeneration}
Ao Fu, Yi Zhou, Tao Zhou, Yi Yang, Bojun Gao, Qun Li, Guobin Wu, and Ling Shao.
\newblock Exploring the interplay between video generation and world models in autonomous driving: A survey, 2024.

\bibitem[Ge et~al.(2022)Ge, Xiao, Xu, Wang, and Itti]{ge2022contributions}
Yunhao Ge, Yao Xiao, Zhi Xu, Xingrui Wang, and Laurent Itti.
\newblock Contributions of shape, texture, and color in visual recognition.
\newblock In \emph{European Conference on Computer Vision}, pages 369--386. Springer, 2022.

\bibitem[Goodfellow et~al.(2014)Goodfellow, Pouget-Abadie, Mirza, Xu, Warde-Farley, Ozair, Courville, and Bengio]{goodfellow2014generative}
Ian Goodfellow, Jean Pouget-Abadie, Mehdi Mirza, Bing Xu, David Warde-Farley, Sherjil Ozair, Aaron Courville, and Yoshua Bengio.
\newblock Generative adversarial nets.
\newblock \emph{Advances in neural information processing systems}, 27, 2014.

\bibitem[Gui et~al.(2023)Gui, Sun, Wen, Tao, and Ye]{gui2020review}
Jie Gui, Zhenan Sun, Yonggang Wen, Dacheng Tao, and Jieping Ye.
\newblock A review on generative adversarial networks: Algorithms, theory, and applications.
\newblock \emph{IEEE Transactions on Knowledge and Data Engineering}, 35\penalty0 (10):\penalty0 1041--1060, 2023.

\bibitem[He et~al.(2024)He, Jiang, Zhang, Ku, Soni, Siu, Chen, Chandra, Jiang, Arulraj, Wang, Do, Ni, Lyu, Narsupalli, Fan, Lyu, Lin, and Chen]{he2024videoscorebuildingautomaticmetrics}
Xuan He, Dongfu Jiang, Ge Zhang, Max Ku, Achint Soni, Sherman Siu, Haonan Chen, Abhranil Chandra, Ziyan Jiang, Aaran Arulraj, Kai Wang, Quy~Duc Do, Yuansheng Ni, Bohan Lyu, Yaswanth Narsupalli, Rongqi Fan, Zhiheng Lyu, Yuchen Lin, and Wenhu Chen.
\newblock Videoscore: Building automatic metrics to simulate fine-grained human feedback for video generation, 2024.

\bibitem[Ho et~al.(2020)Ho, Jain, and Abbeel]{ho2020denoising}
Jonathan Ho, Ajay Jain, and Pieter Abbeel.
\newblock Denoising diffusion probabilistic models.
\newblock \emph{Advances in Neural Information Processing Systems}, 33:\penalty0 6840--6851, 2020.

\bibitem[Ho et~al.(2022{\natexlab{a}})Ho, Chan, Saharia, Whang, Gao, Gritsenko, Kingma, Poole, Norouzi, Fleet, et~al.]{ho2022imagen}
Jonathan Ho, William Chan, Chitwan Saharia, Jay Whang, Ruiqi Gao, Alexey Gritsenko, Diederik~P Kingma, Ben Poole, Mohammad Norouzi, David~J Fleet, et~al.
\newblock Imagen video: High definition video generation with diffusion models.
\newblock \emph{arXiv preprint arXiv:2210.02303}, 2022{\natexlab{a}}.

\bibitem[Ho et~al.(2022{\natexlab{b}})Ho, Salimans, Gritsenko, Chan, Norouzi, and Fleet]{ho2022video}
Jonathan Ho, Tim Salimans, Alexey Gritsenko, William Chan, Mohammad Norouzi, and David~J Fleet.
\newblock Video diffusion models.
\newblock \emph{Advances in Neural Information Processing Systems}, 35:\penalty0 8633--8646, 2022{\natexlab{b}}.

\bibitem[Hu et~al.(2017)Hu, Yang, Liang, Salakhutdinov, and Xing]{hu2017toward}
Zhiting Hu, Zichao Yang, Xiaodan Liang, Ruslan Salakhutdinov, and Eric~P. Xing.
\newblock Toward controlled generation of text.
\newblock In \emph{Proceedings of the 34th International Conference on Machine Learning (ICML)}, pages 1587--1596, 2017.

\bibitem[Huang et~al.(2024)Huang, He, Yu, Zhang, Si, Jiang, Zhang, Wu, Jin, Chanpaisit, et~al.]{huang2024vbench}
Ziqi Huang, Yinan He, Jiashuo Yu, Fan Zhang, Chenyang Si, Yuming Jiang, Yuanhan Zhang, Tianxing Wu, Qingyang Jin, Nattapol Chanpaisit, et~al.
\newblock Vbench: Comprehensive benchmark suite for video generative models.
\newblock In \emph{Proceedings of the IEEE/CVF Conference on Computer Vision and Pattern Recognition}, pages 21807--21818, 2024.

\bibitem[Isola et~al.(2017)Isola, Zhu, Zhou, and Efros]{isola2017image}
Phillip Isola, Jun-Yan Zhu, Tinghui Zhou, and Alexei~A. Efros.
\newblock Image-to-image translation with conditional adversarial networks.
\newblock In \emph{Proceedings of the IEEE Conference on Computer Vision and Pattern Recognition (CVPR)}, pages 1125--1134, 2017.

\bibitem[Karras et~al.(2019)Karras, Laine, and Aila]{karras2019style}
Tero Karras, Samuli Laine, and Timo Aila.
\newblock A style-based generator architecture for generative adversarial networks.
\newblock In \emph{Proceedings of the IEEE/CVF Conference on Computer Vision and Pattern Recognition}, pages 4401--4410, 2019.

\bibitem[Karras et~al.(2021)Karras, Aittala, Laine, Härkönen, Hellsten, Lehtinen, and Aila]{karras2021alias}
Tero Karras, Miika Aittala, Samuli Laine, Erik Härkönen, Janne Hellsten, Jaakko Lehtinen, and Timo Aila.
\newblock Alias-free generative adversarial networks.
\newblock \emph{arXiv preprint arXiv:2106.12423}, 2021.

\bibitem[Khachatryan et~al.(2023)Khachatryan, Movsisyan, Tadevosyan, Henschel, Wang, Navasardyan, and Shi]{khachatryan2023text2videozerotexttoimagediffusionmodels}
Levon Khachatryan, Andranik Movsisyan, Vahram Tadevosyan, Roberto Henschel, Zhangyang Wang, Shant Navasardyan, and Humphrey Shi.
\newblock Text2video-zero: Text-to-image diffusion models are zero-shot video generators, 2023.

\bibitem[Lee et~al.(2023)Lee, Kim, Kim, Lee, and Kang]{lee2023make}
Yoonjin Lee, Minho Kim, Inhan Kim, Sanghun Lee, and Jaegul Kang.
\newblock Make-a-video: Text-to-video generation without text-video data.
\newblock \emph{arXiv preprint arXiv:2303.07637}, 2023.

\bibitem[Li et~al.(2022)Li, Wang, and Yang]{li2022video}
Xinyi Li, Huanyu Wang, and Yi Yang.
\newblock Video transformer network.
\newblock \emph{arXiv preprint arXiv:2102.00719}, 2022.

\bibitem[Li et~al.(2024)Li, Zhou, Zhang, Wei, Hou, and Cheng]{li2024sora}
Xuanyi Li, Daquan Zhou, Chenxu Zhang, Shaodong Wei, Qibin Hou, and Ming-Ming Cheng.
\newblock Sora generates videos with stunning geometrical consistency.
\newblock \emph{arXiv preprint arXiv:2402.17403}, 2024.

\bibitem[Liu et~al.(2024)Liu, Cun, Liu, Wang, Zhang, Chen, Liu, Zeng, Chan, and Shan]{liu2024evalcrafter}
Yaofang Liu, Xiaodong Cun, Xuebo Liu, Xintao Wang, Yong Zhang, Haoxin Chen, Yang Liu, Tieyong Zeng, Raymond Chan, and Ying Shan.
\newblock Evalcrafter: Benchmarking and evaluating large video generation models.
\newblock In \emph{Proceedings of the IEEE/CVF Conference on Computer Vision and Pattern Recognition}, pages 22139--22149, 2024.

\bibitem[Luo et~al.(2023)Luo, Chen, Zhang, Huang, Wang, Shen, Zhao, Zhou, and Tan]{luo2023videofusion}
Zhengxiong Luo, Dayou Chen, Yingya Zhang, Yan Huang, Liang Wang, Yujun Shen, Deli Zhao, Jingren Zhou, and Tieniu Tan.
\newblock Videofusion: Decomposed diffusion models for high-quality video generation.
\newblock \emph{arXiv preprint arXiv:2303.08320}, 2023.

\bibitem[{Pika Labs}(2024)]{pikalabs2024}
{Pika Labs}.
\newblock Pika: Text to video ai, 2024.
\newblock Accessed: 2024-11-14.

\bibitem[Qin et~al.(2024)Qin, Shi, Yu, Wang, Zhou, Li, Yin, Liu, Sheng, Shao, Bai, Ouyang, and Zhang]{qin2024worldsimbenchvideogenerationmodels}
Yiran Qin, Zhelun Shi, Jiwen Yu, Xijun Wang, Enshen Zhou, Lijun Li, Zhenfei Yin, Xihui Liu, Lu Sheng, Jing Shao, Lei Bai, Wanli Ouyang, and Ruimao Zhang.
\newblock Worldsimbench: Towards video generation models as world simulators, 2024.

\bibitem[Radford et~al.(2015)Radford, Metz, and Chintala]{radford2015unsupervised}
Alec Radford, Luke Metz, and Soumith Chintala.
\newblock Unsupervised representation learning with deep convolutional generative adversarial networks.
\newblock \emph{arXiv preprint arXiv:1511.06434}, 2015.

\bibitem[Radford et~al.(2021)Radford, Kim, Hallacy, Ramesh, Goh, Agarwal, Sastry, Askell, Mishkin, Clark, et~al.]{radford2021learning}
Alec Radford, Jong~Wook Kim, Chris Hallacy, Aditya Ramesh, Gabriel Goh, Sandhini Agarwal, Girish Sastry, Amanda Askell, Pamela Mishkin, Jack Clark, et~al.
\newblock Learning transferable visual models from natural language supervision.
\newblock \emph{arXiv preprint arXiv:2103.00020}, 2021.

\bibitem[Ramesh et~al.(2022)Ramesh, Dhariwal, Nichol, Chu, and Chen]{ramesh2022hierarchical}
Aditya Ramesh, Prafulla Dhariwal, Alex Nichol, Casey Chu, and Mark Chen.
\newblock Hierarchical text-conditional image generation with clip latents.
\newblock \emph{arXiv preprint arXiv:2204.06125}, 2022.

\bibitem[Ravi et~al.(2024)Ravi, Gabeur, Hu, Hu, Ryali, Ma, Khedr, R{\"a}dle, Rolland, Gustafson, Mintun, Pan, Alwala, Carion, Wu, Girshick, Doll{\'a}r, and Feichtenhofer]{ravi2024sam2}
Nikhila Ravi, Valentin Gabeur, Yuan-Ting Hu, Ronghang Hu, Chaitanya Ryali, Tengyu Ma, Haitham Khedr, Roman R{\"a}dle, Chloe Rolland, Laura Gustafson, Eric Mintun, Junting Pan, Kalyan~Vasudev Alwala, Nicolas Carion, Chao-Yuan Wu, Ross Girshick, Piotr Doll{\'a}r, and Christoph Feichtenhofer.
\newblock Sam 2: Segment anything in images and videos.
\newblock \emph{arXiv preprint arXiv:2408.00714}, 2024.

\bibitem[Rombach et~al.(2022)Rombach, Blattmann, Lorenz, Esser, and Ommer]{rombach2022high}
Robin Rombach, Andreas Blattmann, Dominik Lorenz, Patrick Esser, and Bj{\"o}rn Ommer.
\newblock High-resolution image synthesis with latent diffusion models.
\newblock In \emph{Proceedings of the IEEE/CVF conference on computer vision and pattern recognition}, pages 10684--10695, 2022.

\bibitem[Rubner et~al.(2000)Rubner, Tomasi, and Guibas]{rubner2000earth}
Yossi Rubner, Carlo Tomasi, and Leonidas~J Guibas.
\newblock The earth mover's distance as a metric for image retrieval.
\newblock \emph{International journal of computer vision}, 40:\penalty0 99--121, 2000.

\bibitem[Salimans et~al.(2016)Salimans, Goodfellow, Zaremba, Cheung, Radford, and Chen]{salimans2016improved}
Tim Salimans, Ian Goodfellow, Wojciech Zaremba, Vicki Cheung, Alec Radford, and Xi Chen.
\newblock Improved techniques for training gans.
\newblock \emph{Advances in neural information processing systems}, 29, 2016.

\bibitem[Soomro et~al.(2012)Soomro, Zamir, and Shah]{soomro2012ucf101}
Khurram Soomro, Amir~Roshan Zamir, and Mubarak Shah.
\newblock Ucf101: A dataset of 101 human actions classes from videos in the wild.
\newblock In \emph{CRCV-TR-12-01}, pages 1--8. University of Central Florida, Center for Research in Computer Vision, 2012.

\bibitem[Tulyakov et~al.(2018)Tulyakov, Liu, Yang, and Kautz]{tulyakov2018mocogan}
Sergey Tulyakov, Ming-Yu Liu, Xiaodong Yang, and Jan Kautz.
\newblock Mocogan: Decomposing motion and content for video generation.
\newblock \emph{IEEE Conference on Computer Vision and Pattern Recognition}, 2018.

\bibitem[Unterthiner et~al.(2018)Unterthiner, Van~Steenkiste, Kurach, Marinier, Michalski, and Gelly]{unterthiner2018towards}
Thomas Unterthiner, Sjoerd Van~Steenkiste, Karol Kurach, Raphael Marinier, Marcin Michalski, and Sylvain Gelly.
\newblock Towards accurate generative models of video: A new metric \& challenges.
\newblock \emph{arXiv preprint arXiv:1812.01717}, 2018.

\bibitem[Vondrick et~al.(2016)Vondrick, Pirsiavash, and Torralba]{vondrick2016generating}
Carl Vondrick, Hamed Pirsiavash, and Antonio Torralba.
\newblock Generating videos with scene dynamics.
\newblock In \emph{Advances in neural information processing systems}, pages 613--621, 2016.

\bibitem[Wang et~al.(2023{\natexlab{a}})Wang, Yuan, Chen, Zhang, Wang, and Zhang]{wang2023modelscopetexttovideotechnicalreport}
Jiuniu Wang, Hangjie Yuan, Dayou Chen, Yingya Zhang, Xiang Wang, and Shiwei Zhang.
\newblock Modelscope text-to-video technical report, 2023{\natexlab{a}}.

\bibitem[Wang and Yang(2024)]{wang2024vidprommillionscalerealpromptgallery}
Wenhao Wang and Yi Yang.
\newblock Vidprom: A million-scale real prompt-gallery dataset for text-to-video diffusion models, 2024.

\bibitem[Wang et~al.(2023{\natexlab{b}})Wang, Zhu, Huang, Chen, Zhu, and Lu]{wang2023drivedreamer}
Xiaofeng Wang, Zheng Zhu, Guan Huang, Xinze Chen, Jiagang Zhu, and Jiwen Lu.
\newblock Drivedreamer: Towards real-world-driven world models for autonomous driving.
\newblock \emph{arXiv preprint arXiv:2309.09777}, 2023{\natexlab{b}}.

\bibitem[Wang et~al.(2024)Wang, He, Fan, Li, Chen, and Zhang]{wang2024driving}
Yuqi Wang, Jiawei He, Lue Fan, Hongxin Li, Yuntao Chen, and Zhaoxiang Zhang.
\newblock Driving into the future: Multiview visual forecasting and planning with world model for autonomous driving.
\newblock In \emph{Proceedings of the IEEE/CVF Conference on Computer Vision and Pattern Recognition}, pages 14749--14759, 2024.

\bibitem[Wang et~al.(2004)Wang, Lu, and Bovik]{wang2004video}
Zhou Wang, Ligang Lu, and Alan~C Bovik.
\newblock Video quality assessment based on structural distortion measurement.
\newblock \emph{Signal processing: Image communication}, 19\penalty0 (2):\penalty0 121--132, 2004.

\bibitem[Wang et~al.(2022)Wang, Cao, Hao, and Sadigh]{wang2022weakly}
Zihan Wang, Zhangjie Cao, Yilun Hao, and Dorsa Sadigh.
\newblock Weakly supervised correspondence learning.
\newblock In \emph{2022 International Conference on Robotics and Automation (ICRA)}, pages 469--476. IEEE, 2022.

\bibitem[Wu et~al.(2023)Wu, Jing, Cheang, Chen, Xu, Li, Liu, Li, and Kong]{wu2023unleashing}
Hongtao Wu, Ya Jing, Chilam Cheang, Guangzeng Chen, Jiafeng Xu, Xinghang Li, Minghuan Liu, Hang Li, and Tao Kong.
\newblock Unleashing large-scale video generative pre-training for visual robot manipulation.
\newblock \emph{arXiv preprint arXiv:2312.13139}, 2023.

\bibitem[Xing et~al.(2024)Xing, Feng, Chen, Dai, Hu, Xu, Wu, and Jiang]{xing2024survey}
Zhen Xing, Qijun Feng, Haoran Chen, Qi Dai, Han Hu, Hang Xu, Zuxuan Wu, and Yu-Gang Jiang.
\newblock A survey on video diffusion models.
\newblock \emph{ACM Computing Surveys}, 57\penalty0 (2):\penalty0 1--42, 2024.

\bibitem[Xu et~al.(2022)Xu, Agrawal, and Dohan]{xu2022videogpt}
Wilson Xu, Aditya Agrawal, and David Dohan.
\newblock Videogpt: Video generation using vq-vae and transformers.
\newblock \emph{arXiv preprint arXiv:2104.10157}, 2022.

\bibitem[Yang et~al.(2023)Yang, Srivastava, and Mandt]{yang2023diffusion}
Ruihan Yang, Prakhar Srivastava, and Stephan Mandt.
\newblock Diffusion probabilistic modeling for video generation.
\newblock \emph{Entropy}, 25\penalty0 (10):\penalty0 1469, 2023.

\bibitem[Yu et~al.(2017)Yu, Zhang, Wang, and Yu]{yu2017seqgan}
Lantao Yu, Weinan Zhang, Jun Wang, and Yong Yu.
\newblock Seqgan: Sequence generative adversarial nets with policy gradient.
\newblock In \emph{Proceedings of the 31st AAAI Conference on Artificial Intelligence (AAAI)}, pages 2852--2858, 2017.

\bibitem[Yu et~al.(2022)Yu, Xie, Wang, Zhang, Liu, Yang, and Fu]{yu2022physics}
Peizhuo Yu, Kang Xie, Ying-Cong Wang, Yuting Zhang, Xiaoxiao Liu, Bo Yang, and Qiang Fu.
\newblock Physics-based human motion estimation and synthesis from videos.
\newblock \emph{arXiv preprint arXiv:2109.09913}, 2022.

\bibitem[Zhang et~al.(2023{\natexlab{a}})Zhang, Rao, and Agrawala]{zhang2023adding}
Lvmin Zhang, Anyi Rao, and Maneesh Agrawala.
\newblock Adding conditional control to text-to-image diffusion models.
\newblock In \emph{Proceedings of the IEEE/CVF International Conference on Computer Vision}, pages 3836--3847, 2023{\natexlab{a}}.

\bibitem[Zhang et~al.(2023{\natexlab{b}})Zhang, Wang, Huang, Tasnim, and Shi]{zhang2023survey}
Tianyi Zhang, Zheng Wang, Jing Huang, Mohiuddin~Muhammad Tasnim, and Wei Shi.
\newblock A survey of diffusion based image generation models: Issues and their solutions.
\newblock \emph{arXiv preprint arXiv:2308.13142}, 2023{\natexlab{b}}.

\bibitem[Zhou et~al.(2023)Zhou, Wang, Yan, Lv, Zhu, and Feng]{zhou2023magicvideoefficientvideogeneration}
Daquan Zhou, Weimin Wang, Hanshu Yan, Weiwei Lv, Yizhe Zhu, and Jiashi Feng.
\newblock Magicvideo: Efficient video generation with latent diffusion models, 2023.

\bibitem[Zhou et~al.(2021)Zhou, Kang, Liu, Borji, and Loy]{zhou2021cocogan}
Xueting Zhou, Bo Kang, Ziwei Liu, Ali Borji, and Chen~Change Loy.
\newblock Cocon: Cooperative-contrastive learning.
\newblock \emph{IEEE Conference on Computer Vision and Pattern Recognition}, 2021.

\end{thebibliography}
}
\clearpage
\setcounter{page}{1}
\maketitlesupplementary

\appendix
\section{Experiment Details}
\label{appendix:experiments}

We utilized the UCF-101 dataset to generate corrupted datasets for evaluation. For benchmarking against state-of-the-art generative models, we randomly sampled 3,000 videos from VidProM \cite{wang2024vidprommillionscalerealpromptgallery}. To ensure consistency, the prompts used for generating the videos were identical across all models.

During score calculation, there were instances where SAM2 failed to produce any segmentation masks. In such cases, we assigned a score of 0 to the corresponding videos. This decision was based on our observations that segmentation failures typically occur for videos of very low quality, particularly those with poor resolution. 

For SIFT-based experiments, the weight tuple (0.3, 0.05, 0.05, 0.6) was applied for the final score calculation, corresponding to the components (color, shape, texture, motion). For SAM2-based experiments, we employed a weight tuple of (0.069, 0.138, 0.092, 0.7), ensuring a balanced contribution of appearance and motion components.

In terms of hardware, we primarily utilized NVIDIA A100 GPUs as accelerators. The inference tests for SAM2 were conducted on a single A100 GPU, using the original SAM2-large (sam2-hiera-large) implementation in a single-process setup. Conversely, SIFT inference tests were performed on a machine equipped with two CPU cores, demonstrating the computational efficiency of the SIFT-based approach.

\section{Point Sampling Algorithms}
\label{appendix:point_sampling}

\begin{table*}[t]
\centering

{\scriptsize

\begin{subtable}[t]{\textwidth}
    \centering
    \begin{tabular}{lcccccc|cccccc}
    \toprule
    \multirow{2}{*}{\textbf{Method}} & \multicolumn{6}{c|}{\textbf{Brightness}} & \multicolumn{6}{c}{\textbf{Gaussian Noise}} \\ 
    \cmidrule(lr){2-7} \cmidrule(lr){8-13}
    & 0 & 1 & 2 & 3 & 4 & 5 
    & 0 & 1 & 2 & 3 & 4 & 5 \\ 
    \midrule
    VAMP-A \(\uparrow\)  & 0.912 & 0.910 & 0.914 & 0.909 & 0.908 & 0.909 & 0.912 & 0.798 & 0.807 & 0.797 & 0.793 & 0.786 \\
    VAMP-M \(\uparrow\)  & 0.578 & 0.569 & 0.571 & 0.564 & 0.557 & 0.559 & 0.578 & 0.508 & 0.518 & 0.526 & 0.507 & 0.499 \\
    \midrule
    VAMP \(\uparrow\)    & 0.712 & 0.706 & 0.708 & 0.702 & 0.697 & 0.699 & 0.712 & 0.624 & 0.634 & 0.634 & 0.621 & 0.614 \\
    \bottomrule
    \end{tabular}
    \label{tab:brightness_gaussiannoise}
\end{subtable}

\vspace{1em} % Add space between the two subtables

\begin{subtable}[t]{\textwidth}
    \centering
    \begin{tabular}{cccccc|cccccc|cccccc}
    \toprule
    \multicolumn{6}{c|}{\textbf{Impulse Noise}} & \multicolumn{6}{c|}{\textbf{Defocus Blur}} & \multicolumn{6}{c}{\textbf{Black Shapes}} \\ 
    \cmidrule(lr){1-6} \cmidrule(lr){7-12} \cmidrule(lr){13-18}
    0 & 1 & 2 & 3 & 4 & 5 & 0 & 1 & 2 & 3 & 4 & 5 & 0 & 1 & 2 & 3 & 4 & 5 \\ 
    \midrule
    0.912 & 0.918 & 0.918 & 0.756 & 0.891 & 0.887 & 0.912 & 0.918 & 0.919 & 0.901 & 0.899 & 0.882 & 0.912 & 0.900 & 0.888 & 0.884 & 0.869 & 0.861 \\
    0.578 & 0.545 & 0.544 & 0.432 & 0.421 & 0.425 & 0.578 & 0.601 & 0.603 & 0.570 & 0.537 & 0.507 & 0.578 & 0.550 & 0.499 & 0.466 & 0.468 & 0.471 \\
    \midrule
    0.712 & 0.694 & 0.694 & 0.562 & 0.609 & 0.610 & 0.712 & 0.728 & 0.729 & 0.703 & 0.682 & 0.657 & 0.712 & 0.690 & 0.655 & 0.634 & 0.628 & 0.627 \\
    \bottomrule
    \end{tabular}
\label{tab:corruption_impulse_defocus_blackshapes}
\end{subtable}

\caption{VAMP score results using the SIFT sampling method across different corruption types (Brightness, Gaussian Noise, Impulse Noise, Defocus Blur, and Black Shapes) at levels 0–5. The table presents scores for three components: VAMP-A (appearance-only), VAMP-M (motion-only), and the combined VAMP score.}
\label{tab:sift_corruption}
}
\end{table*}

\begin{figure*}[!h]
    \centering
    \includegraphics[width=0.95\textwidth]{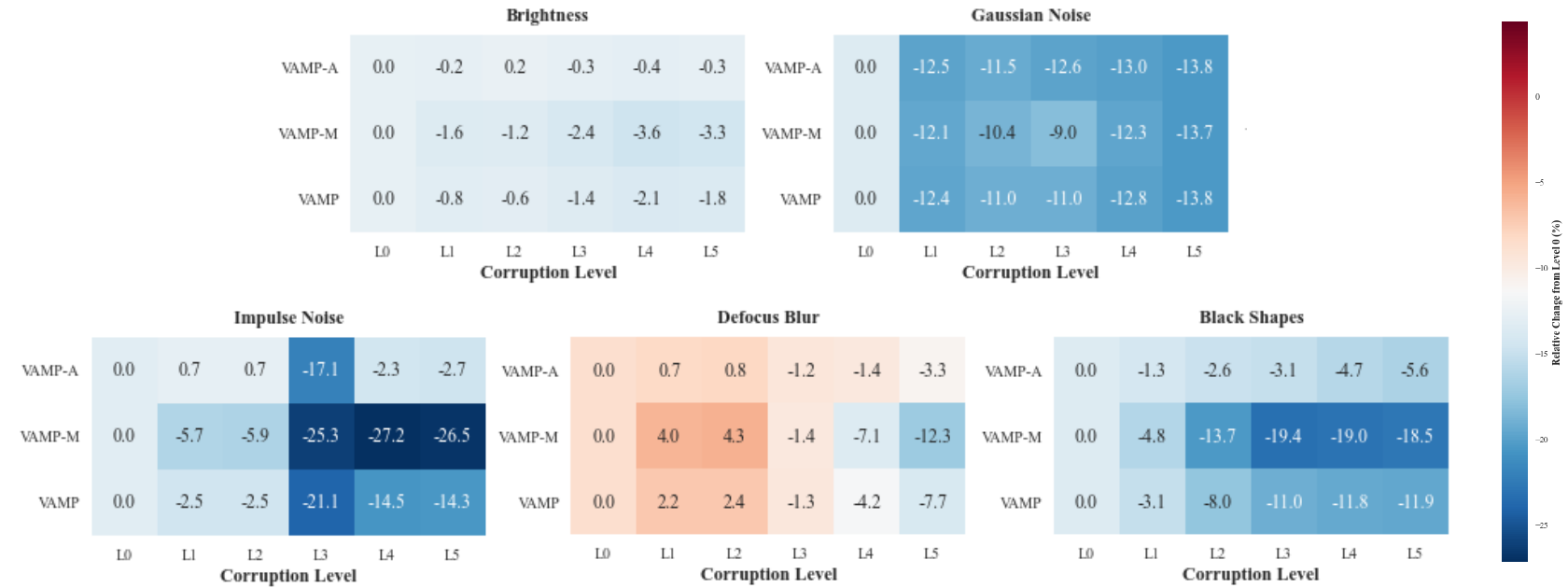}
    \caption{Heatmap of Percentage Changes in VAMP Scores from Level 0 Across Corruption Types Using SIFT Sampling. This figure visualizes the percentage change in VAMP-A (appearance-only), VAMP-M (motion-only), and VAMP (combined) scores across five corruption types: Brightness, Gaussian Noise, Impulse Noise, Defocus Blur, and Black Shapes. Each subplot represents the changes in scores at corruption levels L1 to L5 relative to the baseline at L0.}

\end{figure*}

Effective point sampling is essential for generating relevant masks, which are used for tracking and segmentation of objects in the VAMP pipeline. This section details the four implemented sampling algorithms: Random Sampling, Uniform Grid Sampling, SIFT Sampling and SAM Sampling.

\subsection{Random Sampling}
Random sampling generates a set of points by uniformly sampling coordinates within the dimensions of the input frame. This method is straightforward and computationally efficient, requiring no prior knowledge or feature extraction. The algorithm ensures diversity by randomly selecting \(x\)- and \(y\)-coordinates, effectively covering various regions of the frame.

Although simple, random sampling may lack contextual relevance, as it does not consider the visual content or structure of the frame. However, it is particularly useful as a baseline or fallback method when other algorithms, such as SIFT or SAM, fail to produce reliable points due to insufficient features or challenging input conditions. This method can also be combined with visualization to ensure that the sampled points are uniformly distributed across the frame.

\subsection{Uniform Grid Sampling}
Uniform grid sampling divides the frame into a grid-like structure and generates points at the center of each grid cell. This method ensures consistent spatial coverage, making it ideal for tasks that require uniform distribution of points across the frame. The number of grid cells in the \(x\)- and \(y\)-directions is determined by the specified number of points (\(n_x\) and \(n_y\)), which can be adjusted to balance precision and computational cost.

Uniform grid sampling is particularly suited for scenarios where uniform coverage is essential, such as evaluating global properties of an image or video frame. Unlike random sampling, this method ensures predictable and repeatable point placement. However, it does not adapt to the content of the frame and may miss salient features if they are not aligned with the grid. Its simplicity and consistency make it a reliable choice for structured sampling tasks.

\subsection{SIFT Sampling}
The SIFT sampling method utilizes handcrafted features to identify key points within the input frame. The algorithm detects features by converting the input frame to grayscale and applying the SIFT detector. The resulting key points are returned as sampled points, representing visually salient features based on local gradients and intensity patterns.

For applications requiring clustered feature points, the SIFT-based sampling can be extended with DBSCAN (Density-Based Spatial Clustering of Applications with Noise). The SIFT Cluster algorithm groups detected key points into clusters and extracts their centroids, providing a reduced yet meaningful subset of feature points. This ensures that the sampled points represent the spatial distribution of features in the frame effectively.

The SIFT sampling methods are particularly suited for scenarios where computational efficiency is prioritized, as they rely on traditional feature extraction techniques without the need for pre-trained models.

\subsection{SAM Sampling}
The SAM-based sampling algorithm leverages the Segment Anything Model (SAM) architecture for point generation. The method utilizes the SAM2AutomaticMaskGenerator, a mask-generation module built on top of the SAM2 framework. This approach identifies meaningful regions in the input frame, extracting key points from the mask coordinates. Each point represents a salient feature, ensuring the sampled points are contextually significant. 

This method operates by generating masks for the input frame using the SAM2 model, followed by extracting representative coordinates for each mask. The algorithm supports both CPU and GPU execution, with optimizations such as `torch.autocast` to improve efficiency. SAM sampling excels in scenarios where precise feature localization is critical, as it relies on learned priors from the SAM model to identify features of interest.

\subsection{Efficiency and Performance Tradeoff between SIFT sampling and SAM sampling}

Table \ref{tab:corruption_results} compares the results obtained using the SAM sampling method, while Table \ref{tab:sift_corruption} presents the VAMP scores derived from the SIFT sampling method. A key observation is the tradeoff between efficiency and performance between these two approaches.

\begin{figure*}[!ht]
    \centering
    \includegraphics[width=0.85\textwidth]{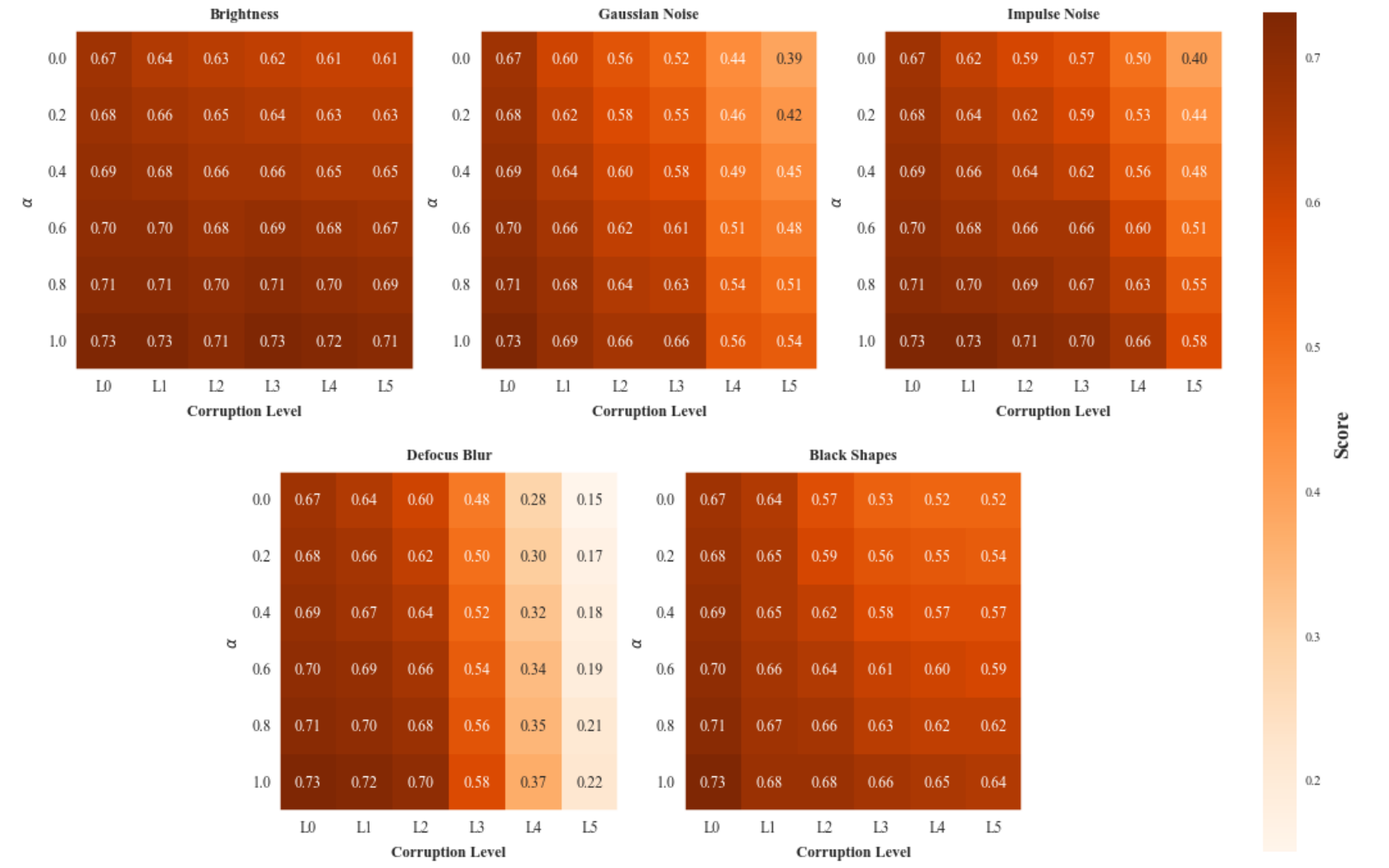}
    \caption{Sensitivity of VAMP Scores to Appearance Weights (\(\alpha\)) Across Corruption Types. 
    Each subplot corresponds to a specific corruption type (Brightness, Gaussian Noise, Impulse Noise, Defocus Blur, and Black Shapes). 
    The x-axis represents the corruption level (0–5), while the y-axis indicates the value of \(\alpha\), the weight of the appearance score in the VAMP metric. 
    The heatmap illustrates how varying \(\alpha\) impacts the VAMP score at different levels of corruption for each type.}
    \label{fig:weight_sensitivity}
\end{figure*}

The SIFT sampling method demonstrates a significant advantage in computational efficiency, with an average sampling time of only 0.095 seconds per frame, compared to 2.69 seconds per frame for SAM2 sampling. This substantial difference makes SIFT sampling a more viable option for applications where rapid inference is required.

\begin{table}[ht]
\centering
\caption{Comparison of Sampling Efficiency Between SIFT and SAM2 Methods}
\label{tab:sampling_efficiency}
\begin{tabular}{lcc}
\toprule
\textbf{Method} & \textbf{Time (s/fr)} & \textbf{Efficiency} \\ 
\midrule
SIFT & 0.095 & High \\ 
SAM2 & 2.69 & Low \\ 
\bottomrule
\end{tabular}
\end{table}

However, this efficiency comes at the cost of performance. The masks generated using SIFT sampling are less representative of the video content compared to those produced by SAM2, leading to less reliable VAMP scores. Specifically, as shown in Table \ref{tab:sift_corruption}, the correlations between corruption levels and VAMP scores are not as ideal, indicating that the SIFT sampling method struggles to capture the impact of corruption types effectively. In contrast, Table \ref{tab:corruption_results} demonstrates that the SAM2 sampling method yields VAMP scores with stronger and more consistent correlations to corruption levels, reflecting its superior ability to represent the overall video quality.

In summary, while SIFT sampling offers notable speed advantages, it sacrifices the representativeness and reliability of the resulting VAMP scores. For tasks that prioritize accurate and meaningful evaluations of video quality, SAM2 sampling is the preferred method despite its higher computational cost. Future work could explore hybrid approaches to balance efficiency and performance.

\section{VAMP Weight Sensitivity Analysis}
\label{appendix:weight_sen}

Weights play a critical role in determining the VAMP score, as they balance the contributions of various components to the overall evaluation. Adjusting these weights emphasizes different characteristics of video quality, such as color, shape, texture, motion, or overall coherence. To better understand the impact of these weights, we present an analysis of VAMP's weight sensitivity. We analyze the sensitivity of the overall VAMP score by varying the weights between the appearance score and motion score.

% \subsection{Sensitivity of Appearance Score Component Weights}

\subsection{Sensitivity of VAMP Component Weights}

\begin{figure}[!ht]
    \centering
    \includegraphics[width=0.85\linewidth]{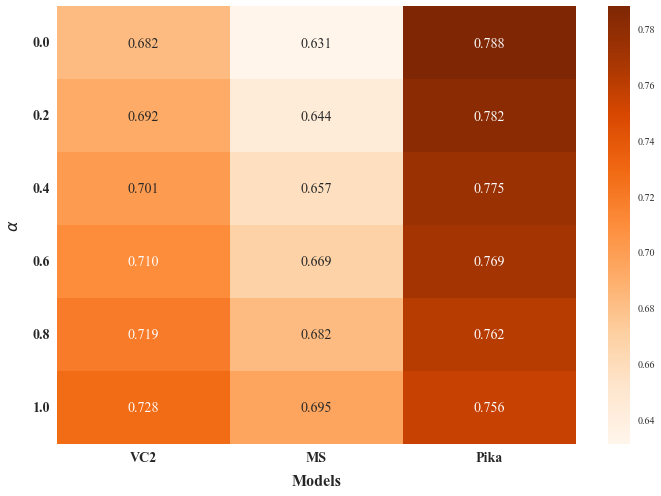} 
    \caption{Impact of \(\alpha\) on VAMP Scores Across Different Video Generation Models. 
    The heatmap compares the VAMP scores for different generative models (VC2, MS, and Pika) as a function of \(\alpha\), the weight of the appearance score.}
    \label{fig:model_performance_alpha}
\end{figure}

The VAMP score (\(S_\text{VAMP}\)) combines appearance (\(S_\text{appearance}\)) and motion (\(S_\text{motion}\)) scores, weighted by \(\alpha\) and \(\beta\), to evaluate video quality. These weights are pivotal to the VAMP metric as they balance the contributions of appearance and motion components, allowing task-specific prioritization and making VAMP adaptable to diverse evaluation scenarios. However, this flexibility necessitates careful weight calibration to ensure meaningful and representative evaluations. To understand the sensitivity of VAMP scores to these weights, we analyze their impact across various corruption types and generative models. Figure \ref{fig:weight_sensitivity} examines the effect of varying the appearance weight (\(\alpha\)) on VAMP scores under five corruption types: Brightness, Gaussian Noise, Impulse Noise, Defocus Blur, and Black Shapes. Figure \ref{fig:model_performance_alpha} explores how \(\alpha\) influences the VAMP scores of three generative models (VC2, MS, and Pika). Below, we discuss the findings from each figure, followed by a summary of key insights.

Figure \ref{fig:weight_sensitivity} illustrates the sensitivity of VAMP scores to \(\alpha\) across different corruption types. For Brightness, VAMP scores remain relatively stable across all corruption levels, suggesting that appearance weight variations have minimal impact. In contrast, Gaussian Noise and Impulse Noise exhibit sharper declines in VAMP scores as corruption levels increase, especially when \(\alpha\) is low, highlighting the importance of motion consistency in these scenarios. Defocus Blur shows a significant drop in VAMP scores at higher corruption levels, particularly when \(\alpha\) is large, indicating that motion plays a crucial role in mitigating its effects. Black Shapes exhibit substantial score reductions at higher corruption levels regardless of \(\alpha\), underscoring the intrinsic difficulty of this corruption type.

Figure \ref{fig:model_performance_alpha} analyzes the impact of \(\alpha\) on the VAMP scores of three generative models, revealing varying sensitivities to appearance and motion components. VC2 demonstrates consistent score improvements as \(\alpha\) increases, indicating a stronger dependence on appearance components. MS (ModelScope) shows moderate sensitivity to \(\alpha\), with a slower increase in scores, reflecting a more balanced reliance on both appearance and motion. Pika consistently outperforms the other models across all \(\alpha\) values, showcasing its robustness in integrating both appearance and motion. Notably, Pika has the smallest sensitivity to \(\alpha\) (0.032 difference), while MS exhibits the largest sensitivity (0.064 difference). These differences indicate that different models prioritize appearance and motion scores differently, with VC2 and MS achieving higher appearance scores and Pika excelling in motion consistency. A more even VAMP score across \(\alpha\) reflects small deviations between appearance and motion scores, while a higher total VAMP value indicates superior overall quality. Pika's high scores and minimal sensitivity suggest it achieves a well-balanced and high-quality output, making it a strong candidate for robust video generation tasks.

In summary, the sensitivity analysis reveals that VAMP scores are highly dependent on the corruption type and the generative model. Corruptions such as Gaussian Noise and Impulse Noise benefit from higher motion weighting (lower \(\alpha\)), while higher appearance weighting (\(\alpha > 0.5\)) is essential for addressing Brightness and Black Shapes. Among the models, Pika demonstrates strong robustness across a wide range of \(\alpha\), making it well-suited for diverse video generation tasks. Future work could focus on adaptive mechanisms to dynamically tune \(\alpha\) based on the corruption type or specific evaluation requirements.

\end{document}